\pdfoutput=1
\documentclass{article}

\usepackage[final]{corl_2026} %

\usepackage[utf8]{inputenc}
\usepackage[T1]{fontenc}
\usepackage{graphicx}
\usepackage{booktabs}
\usepackage{amsfonts}
\usepackage{nicefrac}
\usepackage{microtype}
\usepackage{amsmath,amssymb}
\usepackage{algorithm}
\usepackage{algorithmic}
\usepackage{enumitem}
\usepackage{multirow}
\usepackage{placeins}
\usepackage{wrapfig}
\usepackage{tikz}
\usetikzlibrary{arrows.meta,calc,decorations.pathreplacing,positioning}

\usepackage{xcolor}
\usepackage{hyperref}
\hypersetup{colorlinks=true,urlcolor=blue,citecolor=blue,linkcolor=black}
\hypersetup{
  pdftitle={REACT: Rolling Denoising and Dual Decoupling for Reactive Robot Control with VLA Models},
  pdfauthor={Houlong Xiong, Zhenqi Qiu, Zechen Wang, Suohang Zhang, Yiyu Ren, Wanting Xu, Hongfei Niu, Chengyang He, Ge Sun, Ran Cheng, Qian Zhu},
  pdfkeywords={Reactive Robot Control, Vision-Language-Action Models, Flow Matching}
}

\newcommand{\method}{\textsc{REACT}}

\title{\method{}: Rolling Denoising and Dual Decoupling for Reactive Robot Control with VLA Models}

\author{
  \textbf{Houlong Xiong$^{1,2*}$ \quad
  Zhenqi Qiu$^{2,3*}$ \quad
  Zechen Wang$^{2*}$ \quad
  Suohang Zhang$^{4}$} \\
  \textbf{Yiyu Ren$^{2}$ \quad
  Wanting Xu$^{3}$ \quad
  Hongfei Niu$^{2}$ \quad
  Chengyang He$^{5}$} \\
  \textbf{Ge Sun$^{5}$ \quad
  Ran Cheng$^{2}$ \quad
  Qian Zhu$^{2\dagger}$} \\[6pt]
  \mdseries
  $^{1}$ Shanghai Jiao Tong University \quad
  $^{2}$ PrimeBot \quad
  $^{3}$ ShanghaiTech University \\
  $^{4}$ Zhejiang University \quad
  $^{5}$ National University of Singapore \\
  $^{*}$ Equal contribution \quad $^{\dagger}$ Corresponding author
}

\begin{document}

\maketitle
\vspace{-0.8cm}

\begin{figure}[h]
	\centering
        \includegraphics[
        width=.89\linewidth,
        trim=0.5cm 1cm 3.3cm 1.5cm,
        clip
    ]{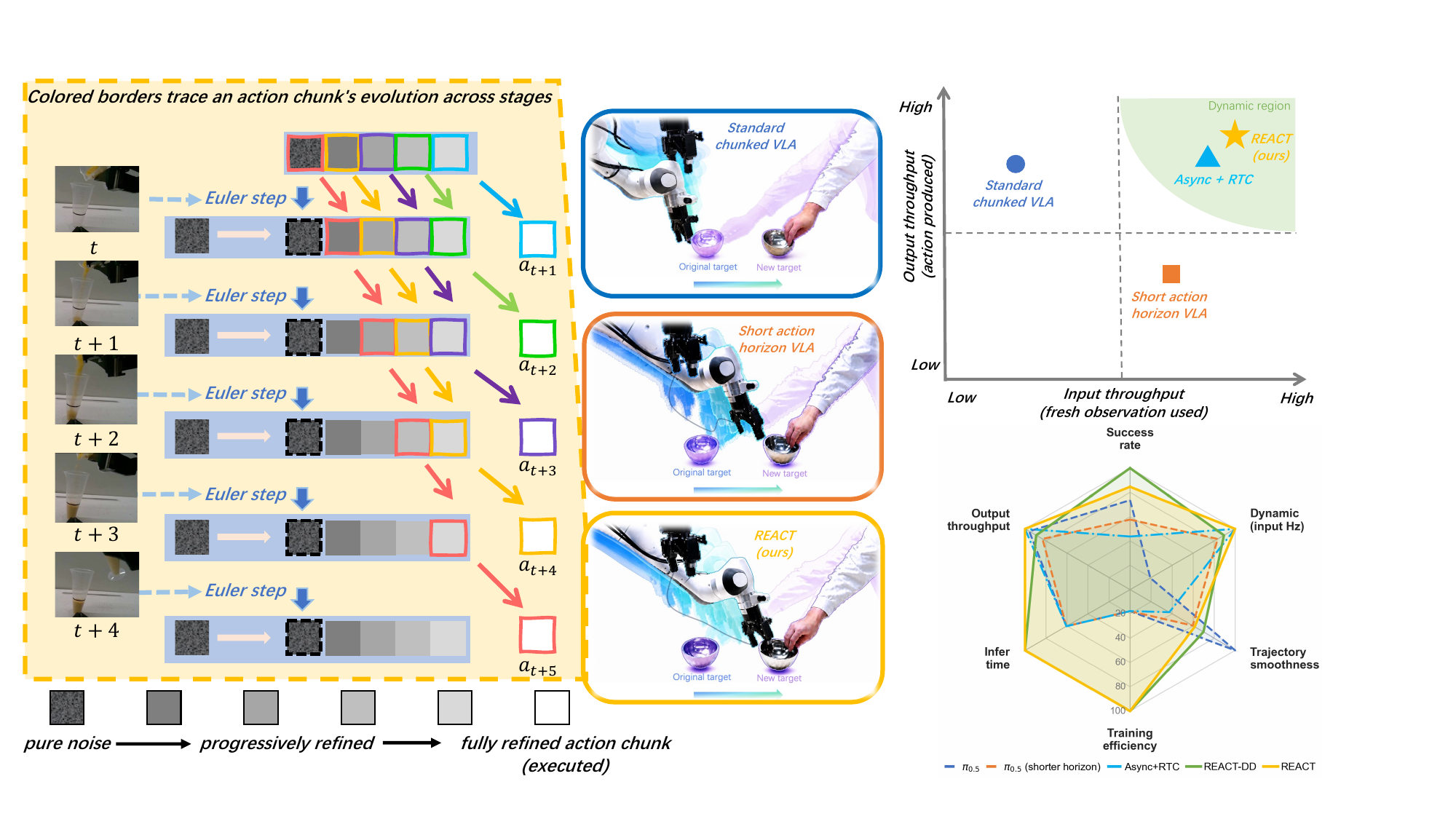}
        \vspace{-0.2cm}
\caption{\textbf{Overview of REACT.}
REACT reformulates flow-based action generation into a receding-horizon rolling denoising process. At each control step, the active horizon is refined conditioned on the latest observation embedding; the oldest refined chunk is executed, while the remaining buffer is shifted forward and replenished with fresh noise. The colored borders track the same action chunk as it is progressively refined across consecutive observations before execution. The right panels summarize throughput and overall performance, showing that REACT reaches the high-input, high-output regime while improving success, reactivity, smoothness, and efficiency.}
\label{fig:front}
\vspace{-0.3cm}
\end{figure}

\begin{abstract}
Flow-based vision-language-action (VLA) models generate action chunks for temporally coherent robot motion, but chunked control creates a fundamental closed-loop trade-off: long chunks provide smooth execution, whereas frequent replanning improves reactivity at the cost of action discontinuities. We introduce \textsc{REACT}, a rolling-denoising framework that makes flow-based VLAs more reactive while preserving long-horizon context. Instead of regenerating entire action chunks from scratch, \textsc{REACT} maintains a persistent action buffer with staggered flow timesteps. At each control step, the full horizon is denoised using the latest observation, the cleanest action block is executed, partially refined future blocks are shifted forward, and fresh noise is appended to the tail. As a result, each executed action block is refined across multiple recent observations before deployment. 
To support real-time control, we further introduce dual decoupling, which separates sensing, VLM encoding, DiT denoising, and action execution, enabling high-frequency observation updates and action streaming under practical compute constraints.
Across the RoboTwin 2.0 simulation benchmark and real-world tasks spanning bimanual manipulation and dynamic control on multiple robot platforms, \textsc{REACT} improves task success and reduces reaction latency while producing smoother trajectories than frequent-replanning and asynchronous baselines.
\end{abstract}


\noindent\textbf{Project page:} \href{https://react-vla.github.io}{react-vla.github.io}
\vspace{0.3em}

\section{Introduction}
Robots in dynamic environments must react quickly to fresh observations while executing smooth, physically stable actions. Vision-language-action (VLA) models have become a promising framework for general-purpose robot control by unifying perception, language grounding, and action generation. %
Many modern VLAs~\cite{brohan2022rt,zitkovich2023rt,kim2024openvla,black2024pi_0,zhao2023learning,chi2025diffusion,intelligence2025pi_} generate continuous action chunks with diffusion or flow-matching objectives.
Action chunking improves temporal consistency, reduces compounding errors, and preserves multimodal action coherence. 
However, it also introduces a closed-loop control limitation: once a chunk is being generated or executed, new observations cannot immediately influence the future actions that remain inside the chunk.

Existing methods improve closed-loop reactivity mainly by reducing or hiding inference latency. Asynchronous inference overlaps prediction with execution, but the next chunk may still be generated from stale observations and can introduce discontinuities when the controller switches between chunks. RTC-style methods mitigate this issue by freezing committed actions and inpainting future actions~\cite{black2026real,black2025training}, while other methods accelerate VLA inference through token optimization, compression, system-level optimization, or faster flow inference~\cite{xu2026vla,pei2025action,yang2026efficientvla,ma2025running,lu2026faster}. We argue that latency alone does not fully characterize dynamic robot control. A closed-loop VLA must simultaneously incorporate fresh observations at high frequency and produce executable actions continuously. We capture these two requirements with \emph{input throughput} and \emph{output throughput}.

This exposes a coherence-freshness trade-off: long chunks preserve smooth, mode-consistent motion but reduce observation freshness, whereas frequent replanning improves reactivity but weakens temporal coherence. To address this trade-off, we propose \textbf{REACT}, a rolling denoising framework for flow-based VLAs. REACT maintains a persistent action buffer with staggered flow times instead of regenerating each chunk from scratch. At each control step, it denoises the active buffer using the latest observation, executes the cleanest front block, shifts partially refined blocks forward, and appends fresh noise. Thus, each executed block is refined across multiple recent observations while the policy retains a long action horizon.

For deployment, we introduce \emph{dual-decoupling}, which separates sensing, VLM encoding, DiT denoising, and action execution without changing model weights, enabling high-frequency observation updates and continuous action streaming. We also introduce matched staircase training to align flow supervision with rolling inference. Across RoboTwin 2.0 simulation, real-world bimanual manipulation on ARX X5 and Franka R3, and targeted Pour Rice and Reaction Game evaluations, REACT improves the success--reactivity trade-off and trajectory smoothness compared with frequent-replanning and asynchronous baselines. Our contributions are:
\begin{itemize}[leftmargin=*, itemsep=1pt, topsep=2pt, parsep=0pt]
    \item We identify the coherence--freshness trade-off in chunked flow-based VLA inference and introduce input/output throughput as system-level measures of closed-loop responsiveness.
    \item We propose \textbf{REACT}, a rolling denoising mechanism that preserves full-horizon action consistency while refining future actions across consecutive observations.
    \item We develop dual-decoupled inference and matched staircase training, enabling high-throughput real-time control with improved success, smoothness, and reactivity.
\end{itemize}

\section{Related Work}

\noindent\textbf{Chunked VLA policies and dynamic reactivity.}
Action chunking is a dominant imitation-learning formulation for visuomotor control, producing action sequences rather than single commands~\cite{lai2025action,zhao2023learning,chi2025diffusion}. It relies on expressive models such as action-chunking transformers, diffusion policies, and flow-matching policies~\cite{zhao2023learning,george2023one,chi2025diffusion,chi2024universal,black2024pi_0,lipman2022flow,he2026demystifying,he2025latent}. Recent work scales them with pretrained vision-language backbones into vision-language-action (VLA) models for generalist robot control~\cite{brohan2022rt,zitkovich2023rt,kim2024openvla,black2024pi_0,intelligence2025pi_,o2024open,cheng2024navila,zhen20243d,zheng2025tracevla,liu2025rdt,he2025falcon}, enabling task and embodiment generalization through larger robot datasets and visual-language pretraining~\cite{o2024open,khazatsky2024droid,walke2023bridgedata,fang2023rh20t,mandlekar2018roboturk,team2025gemini,reuss2025flower}. Yet chunked VLA policies create a coherence--reactivity conflict: longer chunks improve temporal consistency, but observations refresh only at boundaries, so scene changes can leave robots executing stale-context actions.

\noindent\textbf{Efficient inference acceleration and asynchronous inference.}
Prior work improves VLA responsiveness by reducing inference latency or hiding it with asynchronous execution~\cite{yu2025survey,guan2025efficient}. Latency reduction uses compact VLA architectures~\cite{budzianowski2025edgevla,wen2025tinyvla,shukor2025smolvla,chen2025nanovla,williams2025lite,reuss2025flower}, token caching or pruning~\cite{xu2026vla,pei2025action,liu2025vla}, compression~\cite{zhang2026mole,yang2026efficientvla,chen2025rlrc}, and faster generation via speculative decoding, flow acceleration, system optimization, or quantization~\cite{wang2025spec,jia2026action,pertsch2025fast,lu2026faster,ma2025running,dai2025actionflow,wang2025bitvla}. Asynchronous execution predicts the next action chunk during current-chunk execution to remove inter-chunk pauses~\cite{zhao2025vla,jiang2025asyncvla,xie2026dynamicvla,cai2026xiaomi,liu2026learning}, but direct chunk switching can cause observation-state mismatch, discontinuity, performance drops, and trajectory jitter~\cite{liao2025delay,xie2026dynamicvla}. RTC inpaints future actions conditioned on committed prefixes, while training-time RTC, REMAC, and VLASH condition prediction on executed or predicted prefixes~\cite{black2026real,black2025training,wang2026real,tang2025vlash}. These methods reduce or mask latency, but input throughput remains tied to action-execution cadence and may hurt performance or trajectory continuity.

\noindent\textbf{Rolling denoising for generative and robotic policies.}
Rolling denoising extends diffusion sampling from whole-sample refinement to sequential generation, emitting clean variables while appending noisy future ones. Vision and motion methods implement this idea through causal factorization, streaming diffusion, or rolling buffers for frame- or token-level decoding~\cite{ruhe2024rolling,kim2024fifo,kodaira2025streamdit,deng2024causal,zhang2024tedi}. Robotic diffusion policies roll action buffers, relay noise, or model observation-execution delay to avoid full sequence regeneration~\cite{hoeg2024streaming,chen2025responsive,liao2025delay}. However, they target non-VLA policies and do not address VLA inference, where VLM encoding and DiT denoising are coupled bottlenecks. \textsc{REACT} brings rolling denoising to VLA stacks: executed actions exit, noisy actions enter, and active chunks are refined with updated visual-language embeddings. Dual Decoupling separates input processing from action generation, recovering async-like throughput while preserving trajectory continuity.

\section{Methodology}
\label{sec:method}

\subsection{Rolling Denoising for Flow-Based VLAs}
\label{sec:inference}

We consider a flow-based VLA policy~\citep{intelligence2025pi_} that maps an observation $\mathbf{o}_i=(\{I_c\}_{c=1}^{C},\,l,\,s_i)$ to an $H$-step action chunk $\mathbf{A}_i=[\mathbf{a}_i,\ldots,\mathbf{a}_{i+H-1}]\in\mathbb{R}^{H\times D}$ at control iteration $i$.
For a clean action chunk $\mathbf{a}$ and Gaussian noise $\boldsymbol{\epsilon}\sim\mathcal{N}(\mathbf{0},\mathbf{I})$, conditional flow matching~\citep{lipman2022flow} uses the interpolation $\mathbf{x}_{\tau}=\tau\boldsymbol{\epsilon}+(1-\tau)\mathbf{a}$, where $\tau{=}1$ is pure noise and $\tau{=}0$ is clean action, and learns a velocity field $v_{\theta}(\mathbf{x}_{\tau},\tau,\mathbf{o})$ toward $\mathbf{u}_{\tau}=\boldsymbol{\epsilon}-\mathbf{a}$.
Standard inference initializes a noisy chunk and applies $N$ Euler denoising steps conditioned on the same observation.
This keeps all $H$ actions in a shared temporal context, but it ties the entire denoising process to one visual snapshot.
\textsc{REACT} keeps the same $H$-step horizon while rolling the denoising process across consecutive control iterations.
Instead of spending the full denoising budget on one fixed observation, each \textsc{REACT} update performs one Euler step on the rolling buffer under the latest observation, emits the executable front block, shifts the remaining blocks forward, and appends fresh tail noise.

At each iteration, we maintain a rolling action buffer $\mathbf{x}^{(i)}=[B_0^{(i)},B_1^{(i)},\ldots,B_{K-1}^{(i)}]$ with $K$ blocks of length $S$, so $H=KS$.
Block $B_0$ is closest to execution and $B_{K-1}$ represents the far future.
Instead of assigning one flow time to the entire horizon, we assign a per-position staircase time:
\begin{equation}
\label{eq:time_schedule}
    \tau_j = \frac{\lfloor j / S \rfloor + 1}{K}, \quad j = 0, 1, \ldots, H{-}1,
\end{equation}
which yields $\boldsymbol{\tau}=[\underbrace{\tfrac{1}{K},\ldots,\tfrac{1}{K}}_{S},\underbrace{\tfrac{2}{K},\ldots,\tfrac{2}{K}}_{S},\ldots,\underbrace{1,\ldots,1}_{S}]$.
Given the latest $\mathbf{o}_i$, one DiT pass predicts $\mathbf{v}^{(i)}=v_{\theta}(\mathbf{x}^{(i)},\boldsymbol{\tau},\mathbf{o}_i)$, and one Euler update advances every block by one denoising stage:
\begin{equation}
\label{eq:euler_hr}
    \widetilde{\mathbf{x}}^{(i)} = \mathbf{x}^{(i)} - \frac{1}{K}\,\mathbf{v}^{(i)}.
\end{equation}
Thus, block $B_0$ reaches $\tau{=}0$ and becomes executable, while each later block moves to the stage previously occupied by its predecessor.

After the Euler step, \textsc{REACT} executes the updated front block, shifts the remaining blocks left, and appends fresh Gaussian noise to the tail:
\begin{equation}
\label{eq:block_shift}
    \mathbf{x}^{(i+1)} = \left[\, \widetilde{B}_1^{(i)},\, \widetilde{B}_2^{(i)},\, \ldots,\, \widetilde{B}_{K-1}^{(i)},\, \boldsymbol{\epsilon}_{\mathrm{new}} \,\right],
\end{equation}
where $\widetilde{B}_k^{(i)}$ is the $k$-th block after the update and $\boldsymbol{\epsilon}_{\mathrm{new}}\sim\mathcal{N}(\mathbf{0},\mathbf{I})$ reinitializes the far future.
The buffer therefore preserves the same staircase schedule at every iteration, while the contents of each block progressively move toward execution under successive observations.
Steady-state emission begins after $K{-}1$ warm-up iterations; see Appendix~\ref{appendix:inference_config} for the warm-up and inference configuration.

This rolling denoising structure improves observation freshness while preserving the long-horizon mode consistency of action chunking. 
At each block-rate update, \textsc{REACT} refreshes the observation but still denoises the full $H$-step rolling buffer rather than an isolated short-horizon block. 
Thus, adjacent blocks are refined under a shared long-horizon action context, keeping them coupled in the same action mode and smoothing transitions across block boundaries. 
Meanwhile, each newly appended block is refined across $K$ rolling updates as it moves from the noisy tail to the executable front, so the final executed block is shaped by multiple recent observations instead of a single visual snapshot. 
This many-observations-to-one-block refinement progressively disambiguates the multimodal action distribution, encourages the block to converge toward a consistent action-space mode, and reduces the instability caused by independently regenerated short-horizon chunks.
This also distinguishes \textsc{REACT} from RTC-style inpainting~\cite{black2026real,black2025training}: RTC still generates each new chunk from a single observation and stitches it to a committed prefix, which makes the boundary continuous but does not prevent the new suffix from selecting a different action mode. \textsc{REACT} instead preserves one rolling buffer and refines it progressively under consecutive observations, reducing such mode switches before execution.

\subsection{Dual Decoupling for High-Throughput Deployment}
\label{par:dual_decoupling}
Rolling denoising gives \textsc{REACT} the control-quality benefits of rolling multi-observation refinement, but a serialized implementation would behave like a synchronous policy: each update would wait for VLM encoding, DiT denoising, and block execution before the next observation could be used.
We therefore use \textbf{Dual Decoupling}, an inference-time scheduler that leaves the VLA weights unchanged while separating sensing and VLM prefix computation from DiT denoising and action execution.
For block size $S$, the dual-decoupled scheduler selects one observation every $S$ camera periods, yielding a selected-observation rate of $f_{\mathrm{cam}}/S$.
The sensing loop runs at this camera-derived cadence without waiting for block execution, a pool of $M$ VLM workers encodes selected frames asynchronously into a latest-ready cache, and each DiT update consumes the freshest completed embedding together with the rolling buffer.
Here, $M$ denotes the number of VLM worker instances rather than the number of GPUs: workers can be colocated on one GPU when memory permits or distributed across multiple GPUs when per-device throughput is limited.
When the VLM pool is provisioned so that $M/T_{\mathrm{VLM}}\ge f_{\mathrm{cam}}/S$, the visible observation rate and executable-action rate are bounded by
\begin{equation}
\label{eq:dual_decoupling}
    \Phi_{\mathrm{in}}=\min\!\left(\frac{f_{\mathrm{cam}}}{S},\frac{1}{T_{\mathrm{DiT}}}\right),
    \qquad
    \Phi_{\mathrm{out}}=\min\!\left(f_{\mathrm{cam}},\frac{S}{T_{\mathrm{DiT}}}\right).
\end{equation}
The purpose of this decoupling is to recover the input and output throughput that \textsc{REACT} would otherwise lose to serialized synchronization.
For example, when the VLM workers and DiT can sustain the selected observation stream, the system processes observations at $f_{\mathrm{cam}}/S$ and emits $S$ executable actions per update, yielding action throughput at the camera cadence.
The block size $S$ remains a deployment parameter that sets the selected observation interval, but the main role of Dual Decoupling is not flexible rate control; it is to lift full \textsc{REACT} into the asynchronous-throughput regime while preserving the rolling-buffer refinement that improves success and smoothness.
Dual Decoupling trades a small amount of conditioning staleness for this throughput: each DiT update may consume an embedding captured slightly before the update.
We log the embedding age $a=t_{\mathrm{consume}}-t_{\mathrm{capture}}$ for every DiT update; with $S{=}10$ and $M{=}2$ workers, its mean is 233\,ms, its 95th percentile is 383\,ms, and its maximum is 416\,ms, compared with a selected-observation period of 333\,ms, so the conditioning is never more than one selected observation behind.
Out-of-order completions were not observed (0.0\% of updates) and would be harmless in any case, because the latest-ready cache is monotone in capture timestamp: a late worker never regresses the conditioning.
Detailed throughput derivations and the Dual Decoupling timing diagram are provided in Appendix~\ref{appendix:throughput}.

\subsection{Staircase Training for Rolling Denoising}
\label{sec:training}

\paragraph{Staircase training procedure.}
\textsc{REACT} trains the action expert on the same per-position flow times used by rolling inference~\citep{chen2024diffusion}.
Given a demo chunk $(\mathbf{a},\mathbf{o})$, we split the horizon into $K$ blocks, assign $\tau_j$ by Eq.~\eqref{eq:time_schedule}, sample $\boldsymbol{\epsilon}\sim\mathcal{N}(\mathbf{0},\mathbf{I})$, and form
$x_{\tau_j}^{(j)}=\tau_j\epsilon^{(j)}+(1-\tau_j)a^{(j)}$ for each action position.
One forward pass receives the full noisy horizon with per-position time embeddings and predicts all velocities jointly.
The loss supervises the $K$ deterministic denoising stages in one example:
\begin{equation}
\label{eq:multi_t_loss}
    \mathcal{L}_{\text{REACT}} = \frac{1}{K} \sum_{k=0}^{K-1} \frac{1}{SD} \sum_{j \in B_k} \| v_\theta^{(j)} - u_{\tau_j}^{(j)} \|^2,
\end{equation}
where $B_k$ denotes action positions in block $k$.
This objective covers the time-position pairing used by rolling inference: block $k$ is always queried at flow time $(k{+}1)/K$ before shifting one position toward execution, so the training loss supervises the same blockwise noise levels used by the deployed scheduler.
Appendix~\ref{appendix:train_inference_consistency} provides the detailed consistency argument.

\noindent\textbf{Deterministic multi-timestep training.}
Standard flow matching samples a single timestep $\tau \sim p(\tau)$ per training sample,
requiring many epochs to cover the full probability path from $\tau{=}1$ to $\tau{=}0$.
Recent flow-based VLAs similarly densify supervision by evaluating multiple random flow samples per action chunk~\cite{fang2026molmoact2}.
\textsc{REACT} can be viewed as a structured variant of this idea: instead of drawing independent timesteps, it supervises the deterministic set $\{1/K,2/K,\ldots,1\}$ tied to the rolling-buffer positions.
This gives each demonstration chunk coverage over all denoising stages required by the deployed scheduler while preserving the train-inference alignment analyzed in Appendix~\ref{appendix:train_inference_consistency}.
Empirically, this leads to faster convergence and lower open-loop prediction error
(\S\ref{sec:experiments}).

\section{Experiments}
\label{sec:experiments}

Our experiments are designed to answer three questions:
(1) What limits the dynamic performance of existing VLA inference schemes?
(2) How does \textsc{REACT} address these problems?
(3) Beyond addressing these problems, what other benefits does \textsc{REACT} provide?

\subsection{Experimental Setup}
\label{sec:exp_setup}

\noindent\textbf{Policy variants and baselines.}
All methods build on $\pi_{0.5}$~\citep{intelligence2025pi_}, a flow-based VLA, to isolate the effect of the closed-loop inference schedule. We compare against $\pi_{0.5}$ variants that expose the coherence--reactivity trade-off in chunked control: \textbf{$\pi_{0.5}$ H50E50} trains with horizon $H{=}50$ and executes the full 50-action chunk before re-planning; \textbf{$\pi_{0.5}$ H50E10} keeps the same training horizon but executes only the first 10 actions; \textbf{$\pi_{0.5}$ H10E10} trains and executes with a shorter $H{=}10$ horizon; and \textbf{Async + Training RTC} uses asynchronous inference with a 10-action inference interval and training-time Real-Time Control~\citep{black2025training}, conditioning predictions on executed actions while keeping the original training recipe. We denote Dual Decoupling as DD. Our variants are \textbf{\textsc{REACT} w/o DD}, which isolates rolling denoising, and full \textbf{\textsc{REACT}}, which adds DD; both use block size $S{=}10$ and $K{=}5$ blocks. We choose this setting based on a block-granularity ablation over $S$ and $K$ reported in Appendix~\ref{appendix:ablation}. Model architecture and action-representation details are provided in Appendix~\ref{appendix:Architecture}.

\begin{figure}[H]
    \centering
    \includegraphics[width=\linewidth]{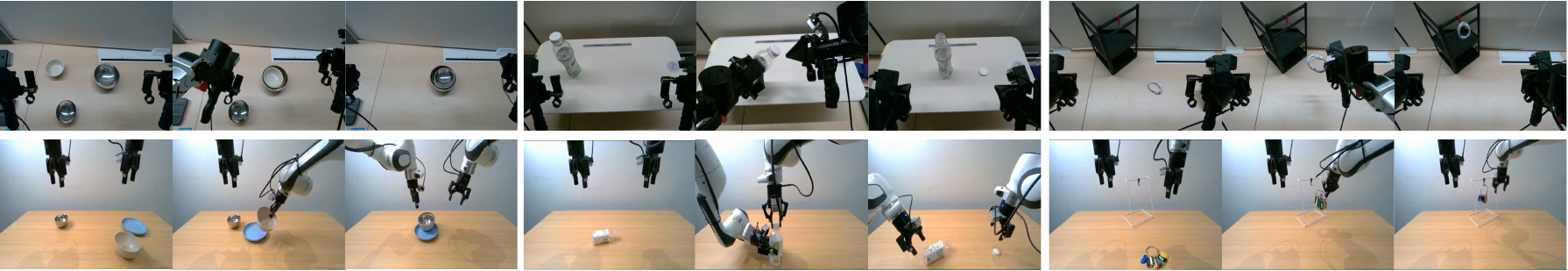}
    \vspace{-1.0em}
    \caption{\textbf{Real-world general manipulation tasks.} ARX X5 (top) and Franka R3 (bottom) perform two shared tasks, Bowl Stacking by Size and Bottle Cap Unscrewing, plus robot-specific hanging tasks: Cable Hanging and Keyring Hanging, respectively.}
    \label{fig:real_tasks}
    \vspace{-1.0em}
\end{figure}

\noindent\textbf{Benchmarks, platforms, and evaluation protocol.}
We evaluate in RoboTwin 2.0 simulation~\citep{chen2025robotwin} and on two real bimanual platforms, \textbf{ARX Robotics X5} (ARX X5) and \textbf{Franka Research 3} (Franka R3). For simulation, we use seven representative tasks under Clean and Randomized conditions (Table~\ref{tab:sim_results}), train each policy with 50 expert demonstrations per task, and evaluate 100 rollouts per task-condition pair. For real-world evaluation, we use six platform-task pairs (Figure~\ref{fig:real_tasks}) spanning size-based reasoning, bimanual coordination, and deformable or articulated object manipulation, with 30 trials per task under matched randomized object placements across methods. We report \textbf{Success Rate (\%)} as the primary task-completion metric and \textbf{Trajectory Smoothness} as a motion-quality metric, measured by mean joint-trajectory jerk over successful episodes only. Specialized tasks additionally use task-specific metrics introduced in \S\ref{sec:specialized_results}. Full task lists, benchmark protocol, hardware specifications, and metric definitions are provided in Appendices~\ref{appendix:hardware}, \ref{appendix:benchmark_protocol}, and~\ref{appendix:evaluation_metrics}.

\subsection{General Task Results}
\label{sec:general_results}

\FloatBarrier

We first evaluate manipulation performance in simulation and on real robots, following the protocol in \S\ref{sec:exp_setup}. This broad evaluation compares existing VLA inference schemes and \textsc{REACT} across task success and trajectory smoothness, providing the main evidence for the first two questions.

\begin{table}[tbp]
\vspace{-0.5em}
\centering
\caption{\textbf{Simulation results on RoboTwin 2.0 benchmark.} Success rate (\%) under clean and randomized settings. Best in \textbf{bold}, second best \underline{underlined}.}
\label{tab:sim_results}
\resizebox{\linewidth}{!}{%
\begin{tabular}{@{}l cc cc cc cc cc cc@{}}
\toprule
& \multicolumn{2}{c}{$\pi_{0.5}$ H50E50} & \multicolumn{2}{c}{$\pi_{0.5}$ H50E10} & \multicolumn{2}{c}{$\pi_{0.5}$ H10E10} & \multicolumn{2}{c}{Async+RTC} & \multicolumn{2}{c}{\textbf{\textsc{REACT} w/o DD}} & \multicolumn{2}{c}{\textbf{\textsc{REACT}}} \\
\cmidrule(lr){2-3} \cmidrule(lr){4-5} \cmidrule(lr){6-7} \cmidrule(lr){8-9} \cmidrule(lr){10-11} \cmidrule(lr){12-13}
Task & Clean & Rand & Clean & Rand & Clean & Rand & Clean & Rand & Clean & Rand & Clean & Rand \\
\midrule
beat\_block\_hammer    & \textbf{17} & 1 & 8 & 0 & 0 & 0 & 13 & \textbf{4} & \underline{15} & \underline{2} & 14 & \textbf{4} \\
click\_bell            & 61 & 6 & \underline{71} & 18 & 35 & 0 & 64 & 8 & \textbf{89} & \textbf{32} & 61 & \underline{22} \\
move\_playingcard\_away & 59 & \textbf{26} & \textbf{71} & 13 & 7 & 1 & 64 & 11 & \textbf{71} & \underline{25} & \underline{66} & 22 \\
pick\_diverse\_bottles & 20 & 3 & 9 & 0 & 0 & 1 & 10 & \underline{6} & \underline{27} & \textbf{9} & \textbf{33} & \underline{6} \\
press\_stapler         & 58 & 23 & 22 & 3 & 9 & 4 & 37 & 13 & \textbf{67} & \textbf{42} & \underline{63} & \underline{39} \\
rotate\_qrcode         & 30 & 3 & 21 & 3 & 4 & 1 & 12 & \underline{8} & \textbf{52} & 6 & \underline{46} & \textbf{11} \\
turn\_switch           & 17 & 18 & \textbf{34} & 24 & 21 & 10 & \underline{33} & 24 & 27 & \underline{30} & 31 & \textbf{35} \\
\midrule
\textbf{Average}       & 37.43 & 11.43 & 33.71 & 8.71 & 10.86 & 2.43 & 33.29 & 10.57 & \textbf{49.71} & \textbf{20.86} & \underline{44.86} & \underline{19.86} \\
\bottomrule
\end{tabular}%
}
\vspace{-1.0em}
\end{table}

\begin{table}[tbp]
\centering
\caption{\textbf{Real-world results on general manipulation tasks.} Success rate (\%) across ARX X5 and Franka platforms. Best in \textbf{bold}, second best \underline{underlined}.}
\label{tab:real_results}
\resizebox{\linewidth}{!}{%
\begin{tabular}{@{}l l cccccc@{}}
\toprule
Robot & Task & $\pi_{0.5}$ H50E50 & $\pi_{0.5}$ H50E10 & $\pi_{0.5}$ H10E10 & Async+RTC & \textsc{REACT} w/o DD & \textsc{REACT} \\
\midrule
\multirow{3}{*}{ARX X5}
    & Bowl Stacking by Size     & 73 & 66 & 70 & 36 & \textbf{86} & \underline{76} \\
    & Bottle Cap Unscrewing     & 36 & 0 & 0 & 0 & \textbf{56} & \underline{47} \\
    & Cable Hanging             & 56 & \underline{66} & 56 & 3 & 63 & \textbf{80} \\
\cmidrule{2-8}
& \textit{ARX X5 Avg.}          & 55.0 & 44.0 & 42.0 & 13.0 & \textbf{68.3} & \underline{67.7} \\
\midrule
\multirow{3}{*}{Franka}
    & Bowl Stacking by Size     & 60 & 50 & 0 & 47 & \textbf{66} & \underline{63} \\
    & Bottle Cap Unscrewing     & 57 & 0 & 0 & 3 & \underline{63} & \textbf{70} \\
    & Keyring Hanging           & \textbf{70} & 33 & 0 & 17 & \underline{60} & 53 \\
\cmidrule{2-8}
& \textit{Franka Avg.}          & \underline{62.3} & 27.8 & 0.0 & 22.3 & \textbf{63.0} & 62.0 \\
\midrule
\multicolumn{2}{l}{\textbf{Overall Avg.}} & 58.7 & 35.9 & 21.0 & 17.7 & \textbf{65.7} & \underline{64.8} \\
\bottomrule
\end{tabular}%
}
\vspace{-1.0em}
\end{table}

\begin{figure}[tbp]
    \centering
    \includegraphics[width=\linewidth]{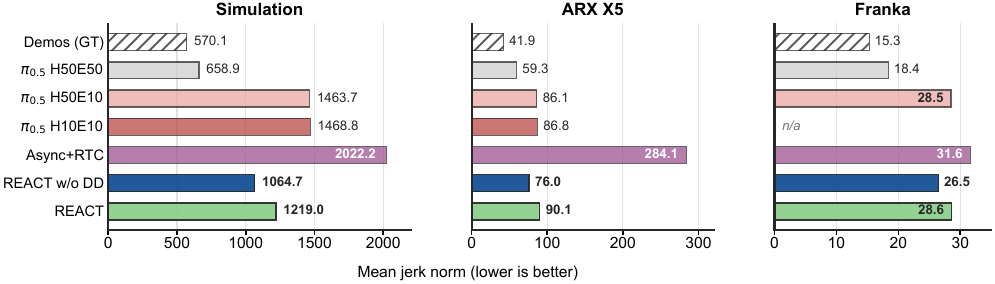}
    \vspace{-1.0em}
    \caption{\textbf{Trajectory smoothness comparison.} Mean jerk norm averaged over successful episodes in simulation, ARX X5, and Franka settings. Demos (GT) applies the same metric to the imitation-learning demonstrations as a reference. Full per-task values are reported in Appendix~\ref{appendix:smoothness_full}.}
    \label{fig:smoothness_comparison}
    \vspace{-1.5em}
\end{figure}

Table~\ref{tab:sim_results}, Table~\ref{tab:real_results} and Figure~\ref{fig:smoothness_comparison} show that dynamic performance of existing chunked VLA inference is constrained by a coherence--freshness conflict, which cannot be resolved by simply increasing replanning frequency. H50E50 executes the full 50-action chunk, preserving the strongest temporal coherence and achieving the lowest simulation jerk among policies (658.9), closest to the demonstrations themselves (570.1), but this long open-loop execution makes the policy less responsive to dynamic scene changes. H50E10 and H10E10 are two naive strategies that improve the observation-refresh rate, but both reduce success and increase jerk. H50E10 drops from $37.43\%$/$11.43\%$ to $33.71\%$/$8.71\%$ on Clean/Randomized simulation, decreases real-world success from $58.7\%$ to $35.9\%$, and increases simulation jerk to 1463.7. H10E10 degrades further because it also lacks long-horizon context during training, reaching only $10.86\%$/$2.43\%$ in simulation and $21.0\%$ in real-world tasks. These results indicate that short-horizon replanning improves nominal observation freshness, but weakens long-horizon action coherence, amplifies action-mode drift across replanning boundaries, and can cause repeated retries and OOD trapping when the policy fails to make state-changing progress. Appendix~\ref{appendix:failure_analysis} provides a detailed failure analysis.

\textsc{REACT} w/o DD isolates the effect of rolling denoising: it preserves the full $H{=}50$ rolling buffer and refines future action blocks under multiple observations selected at block intervals, rather than independently sampling short chunks. This preserves long-horizon action consistency and further narrows the action distribution, reducing multimodality. Compared with the three synchronous baselines, \textsc{REACT} w/o DD achieves the highest average simulation success, with $49.71\%$ on Clean and $20.86\%$ on Randomized settings; it also reaches the highest real-world overall success at $65.7\%$. Although \textsc{REACT} w/o DD is not as smooth as the fully open-loop H50E50 baseline, its simulation jerk is 1064.7, clearly lower than the more reactive synchronous baselines H50E10 (1463.7) and H10E10 (1468.8). This result shows that rolling denoising improves observation freshness while preserving the action consistency and smoothness of long-horizon chunked control.

Full \textsc{REACT} adds Dual Decoupling to raise the throughput of rolling denoising to the same level as asynchronous execution, as shown in Table~\ref{tab:throughput_values}, making it more suitable for real-time deployment. Results show that full \textsc{REACT} substantially outperforms Async+RTC: it improves Clean/Randomized simulation success from $33.29\%$/$10.57\%$ to $44.86\%$/$19.86\%$, and real-world success from $17.7\%$ to $64.8\%$. Meanwhile, compared with async execution using RTC, it reduces jerk from 2022.2 to 1219.0 in simulation, from 284.1 to 90.1 on ARX X5, and from 31.6 to 28.6 on Franka. This shows that \textsc{REACT} with Dual Decoupling achieves the same throughput as async execution without causing the large success degradation and trajectory jitter observed in async execution. These differences exceed evaluation-sampling noise: a two-proportion $z$-test gives $z{=}4.44$, $4.84$, and $9.10$ for clean simulation, randomized simulation, and real-world evaluation, respectively (all $p{<}10^{-5}$). Adding Dual Decoupling does not discard the cross-block modal consistency and smoothness brought by rolling denoising. At the same time, the small gap between \textsc{REACT} and \textsc{REACT} w/o DD is an expected deployment trade-off: the latest-ready embedding cache can introduce residual observation--state mismatch, which rolling denoising greatly mitigates but cannot fully eliminate. On these general tasks, where reactivity matters less, this staleness can leave full \textsc{REACT} slightly below \textsc{REACT} w/o DD; on the specialized dynamic tasks in \S\ref{sec:specialized_results}, the throughput benefit dominates and the ordering reverses.

\FloatBarrier

\subsection{Specialized Task Results}
\label{sec:specialized_results}

The general results above suggest that dynamic manipulation depends on both observation freshness and temporally coherent execution. We therefore further evaluate two specialized tasks that isolate these requirements: closed-loop control under continuously changing physical states and high-frequency reaction to sudden visual events.

\vspace{-0.2em}
\begin{figure}[H]
    \centering
    \begin{minipage}[t]{0.495\linewidth}
        \vspace{0pt}
        \centering
        \includegraphics[width=\linewidth,trim=100pt 235pt 100pt 95pt,clip]{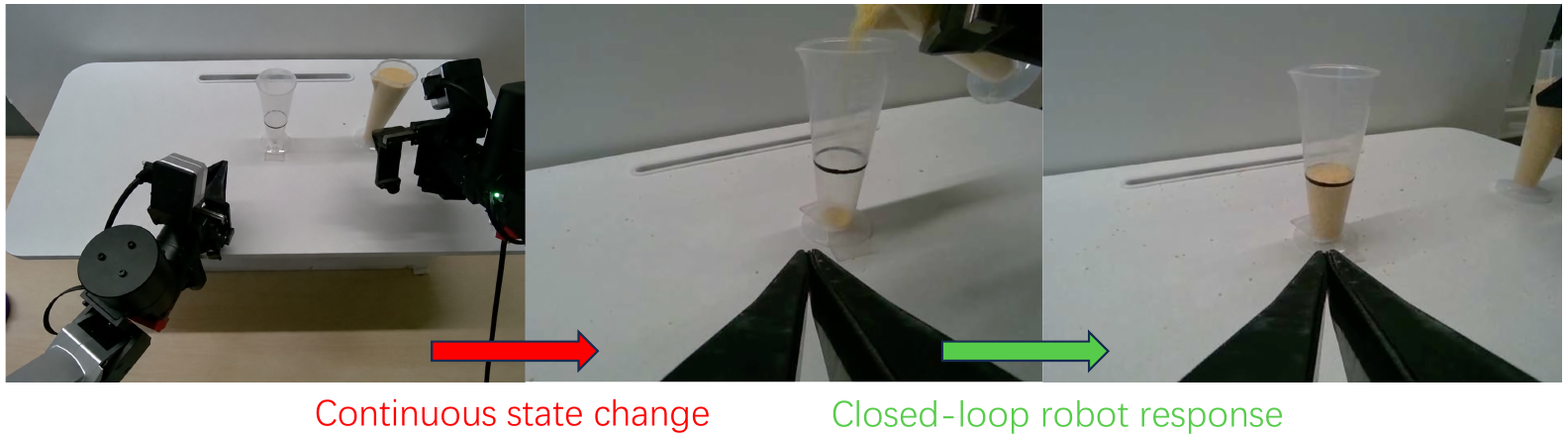}\par
        {\small (a) \textbf{Pour Rice on ARX X5}}\par
    \end{minipage}
    \hfill
    \begin{minipage}[t]{0.495\linewidth}
        \vspace{0pt}
        \centering
        \includegraphics[width=\linewidth]{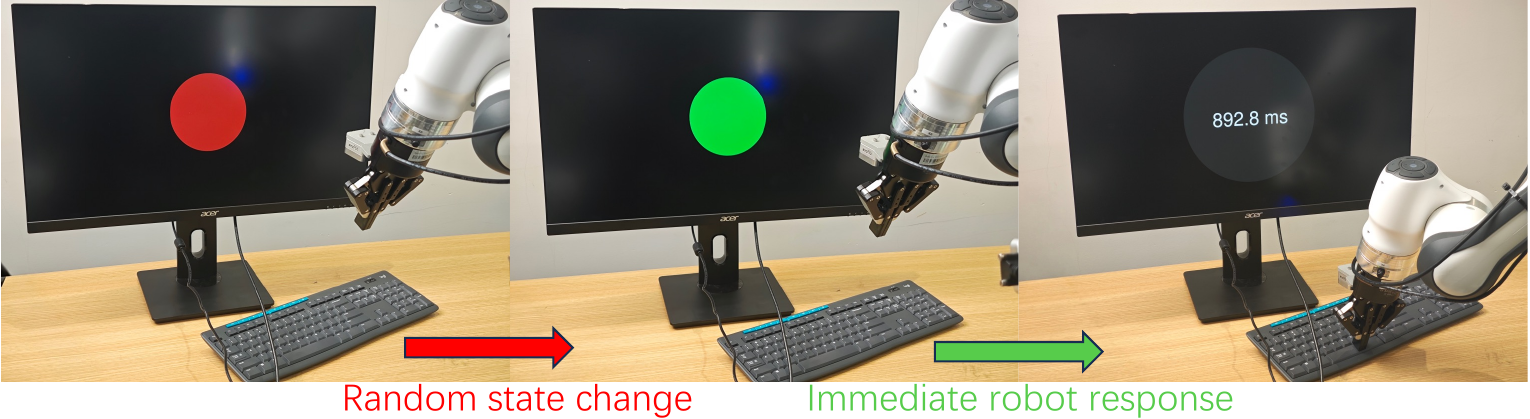}\par
        {\small (b) \textbf{Reaction Game on Franka R3}}\par
    \end{minipage}
    \vspace{-0.5em}
    \caption{\textbf{Specialized dynamic closed-loop control tasks.} (a) \textbf{Pour Rice} requires the robot to pour rice into a vessel and stop at the target state, testing whether the policy can react to continuously changing flow while maintaining smooth motion. (b) \textbf{Reaction Game} requires the robot to click the mouse when the on-screen signal turns green, directly measuring visual-to-motor reaction latency.}
    \label{fig:specialized_tasks}
\end{figure}
\vspace{-1.3em}

\begin{table}[H]
\centering
\caption{\textbf{Specialized task results.} Pour Rice reports success rate (\%) (higher is better). Reaction Game reports mean $\pm$ standard deviation latency in milliseconds (lower is better).}
\label{tab:specialized_results}
\resizebox{\linewidth}{!}{%
\begin{tabular}{@{}l l ccccccc@{}}
\toprule
Task & Metric & $\pi_{0.5}$ H50E50 & $\pi_{0.5}$ H50E10 & $\pi_{0.5}$ H10E10 & Async+RTC & \textsc{REACT} w/o DD & \textsc{REACT} & Human Avg. \\
\midrule
Pour Rice (ARX X5) & Success (\%) $\uparrow$ & 3 & 33 & 0 & 0 & \underline{57} & \textbf{63} & -- \\
Reaction Game (Franka R3) & Latency (ms) $\downarrow$ & 1505 $\pm$ 502 & 768 $\pm$ 113 & 772 $\pm$ 119 & \underline{747 $\pm$ 123} & 754 $\pm$ 127 & \textbf{734 $\pm$ 94} & 705 $\pm$ 73 \\
\bottomrule
\end{tabular}%
}
\vspace{-1.3em}
\end{table}

\textbf{Pour Rice} and \textbf{Reaction Game} expose complementary dynamic-control demands: continuous physical-state changes and sudden visual events. In Pour Rice, H50E50 reacts too late to the rising rice level and over-pours, revealing the cost of stale observations in a long open-loop segment. H50E10 and H10E10 refresh observations more often but suffer short-horizon failures such as OOD pauses and repeated re-grasp attempts, while Async+RTC is more reactive but often spills due to jerky stitched trajectories. Full \textsc{REACT} achieves the highest Pour Rice success rate ($63\%$) by combining fast closed-loop correction with smoother action evolution, avoiding both stale over-pouring and short-horizon mode oscillations. Reaction Game confirms the same advantage under direct latency stress: H50E50 remains dominated by observation staleness, reaching 1505\,ms, whereas full \textsc{REACT} reaches the lowest robot latency at 734\,ms, ahead of Async+RTC (747\,ms) and only 29\,ms above human teleoperation (705\,ms). Together, these tasks show that \textsc{REACT} reacts quickly while preserving the smooth, temporally coherent actions needed for physical execution.

\FloatBarrier

\subsection{Training Efficiency Analysis}
\label{sec:training_efficiency}

\begin{figure}[H]
    \vspace{-1.6em}
    \centering
    \begin{minipage}[t]{0.56\linewidth}
        \centering
        \includegraphics[width=\linewidth]{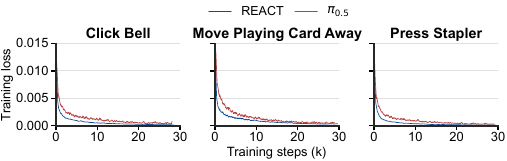}\par
        {\small (a) \textbf{Training loss over optimization steps}}\par
    \end{minipage}
    \hfill
    \begin{minipage}[t]{0.39\linewidth}
        \centering
        \includegraphics[width=0.95\linewidth]{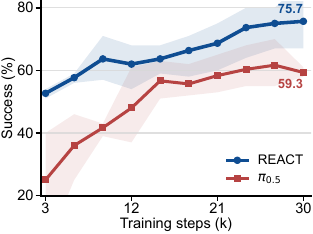}\par
        {\small (b) \textbf{Success vs. training steps}}\par
    \end{minipage}
    \vspace{-0.3em}
    \caption{\textbf{Training efficiency test.} (a) Training loss over optimization steps. (b) Average closed-loop success rate as a function of training steps; shaded bands show the task-wise range.}
    \label{fig:training_efficiency_summary}
    \label{fig:training_loss}
    \label{fig:success_vs_training_steps}
    \vspace{-1.0em}
\end{figure}
\vspace{-0.3em}

Beyond the dynamic-control improvements above, we evaluate whether the training recipe in \S\ref{sec:training} also improves training efficiency. The recipe supervises $K$ deterministic denoising stages in each forward pass, so it learns the deployed action-flow schedule with fewer optimization steps than standard single-timestep flow training. We test this on three representative RoboTwin tasks, Click Bell, Move Playing Card Away, and Press Stapler, using two main measurements: training loss over optimization steps and closed-loop task success as a function of training steps. Appendix~\ref{appendix:open_loop_analysis} additionally reports open-loop action prediction on the \texttt{demo\_randomized} OOD dataset.

Figure~\ref{fig:training_efficiency_summary}(a) shows that \textsc{REACT} lowers training loss faster than the $\pi_{0.5}$ baseline on all three tasks. This trend supports the optimization-efficiency interpretation of the staircase objective: by supervising the same deterministic denoising stages used at deployment, the model receives denser guidance for the action-flow schedule it executes. Figure~\ref{fig:training_efficiency_summary}(b) shows that this faster optimization also translates to closed-loop success: \textsc{REACT} reaches $52.7\%$ average success after only 3k steps versus $25.0\%$ for the $\pi_{0.5}$ baseline and matches the baseline's 30k-step performance using only 15k steps. Thus, beyond improving inference-time dynamic control, \textsc{REACT} also makes training more efficient in terms of optimization steps.

\section{Conclusion}
\label{sec:conclusion}

This paper presented \textsc{REACT}, a rolling-denoising framework for making flow-based VLA policies more reactive while preserving the temporal coherence of long-horizon action chunks. Rather than treating closed-loop control as a choice between stale long chunks and unstable frequent replanning, \textsc{REACT} maintains a persistent action buffer in which future action blocks are progressively refined under successive observations before execution. Together with Dual Decoupling and matched staircase training, this design improves the success--smoothness--reactivity trade-off on RoboTwin 2.0 and in real-world tasks, including bimanual manipulation and dynamic control. These results suggest that real-time VLA control should be evaluated not only by model inference latency, but also by how effectively fresh observations are incorporated into coherent executable actions.

\section{Limitations}
\label{sec:limitations}

A primary limitation is that we validate \textsc{REACT} only on the flow-based VLA backbone $\pi_{0.5}$~\citep{intelligence2025pi_}, leaving adaptation to other VLA models untested. This limits empirical evidence for cross-model compatibility, although \textsc{REACT} only assumes a dual-system interface for perception-language-conditioned action generation and is not tied to model scale or a specific backbone. It could therefore apply to dual-system VLAs such as GR00T N1~\citep{bjorck2025gr00t} and X-VLA~\citep{zheng2025x}; we plan to validate \textsc{REACT} on more dual-system VLAs and to extend it to dual-system video-action models such as mimic-video~\citep{pai2025mimic} and LingBot-VA~\citep{li2026causal}.
A second limitation is sensitivity to the rolling schedule $(K,S)$: each policy is trained for one fixed schedule, and we neither evaluate nor claim zero-shot switching between schedules.
In practice, $S$ can be chosen from the required reaction timescale and compute budget, and $K$ set to the largest integer with $KS\le50$, the pretrained horizon of $\pi_{0.5}$; the same $(K,S)=(5,10)$ was used in all main experiments without per-task tuning (Appendix~\ref{appendix:ablation}).
Future work could train a schedule-conditioned policy by sampling valid block sizes or jointly supervising multiple $(K,S)$ configurations on the same VLM context, similar to the multiple-flow-sample design of MolmoAct2~\citep{fang2026molmoact2}; a state-conditioned controller could then select among supported schedules according to the complexity and uncertainty of each task phase.
Finally, we did not train independent policy seeds, so our statistical comparisons reflect evaluation-rollout sampling noise rather than training-seed robustness.

\bibliography{references}

\appendix

\section{Failure Case Analysis}
\label{appendix:failure_analysis}

We expand on the first analysis in \S\ref{sec:general_results} to explain why higher nominal dynamics do not necessarily produce better closed-loop manipulation.
Across the general manipulation tasks and the two specialized dynamic tasks, the failed baselines show three recurring mechanisms: long open-loop execution remains successful and smooth in standard settings but becomes stale under dynamic evaluation, shortened horizons improve observation freshness while weakening action-distribution support, and asynchronous execution improves output streaming while introducing trajectory stitching artifacts.
These mechanisms are visible not only in qualitative rollouts, but also in the paired success, latency, and jerk measurements in Tables~\ref{tab:sim_results}--\ref{tab:specialized_results} and Table~\ref{tab:smoothness_full}.

\begin{figure}[tbp]
    \centering
    \includegraphics[width=\linewidth,trim=0 360pt 0 0,clip]{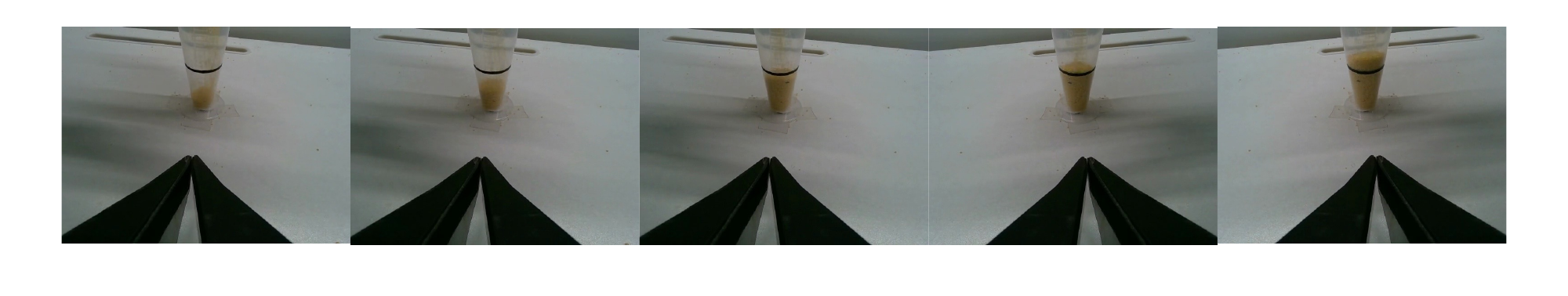}
    \caption{\textbf{Over-pouring under long open-loop execution in Pour Rice.}
    Consecutive frames from the H50E50 baseline show the policy continuing the pouring motion even after the rice level exceeds the target mark, illustrating how stale open-loop execution can overshoot the target state before the next observation update.}
    \label{fig:pour_rice_over_pi}
    \vspace{-0.8em}
\end{figure}

\begin{figure}[tbp]
    \centering
    \includegraphics[width=\linewidth]{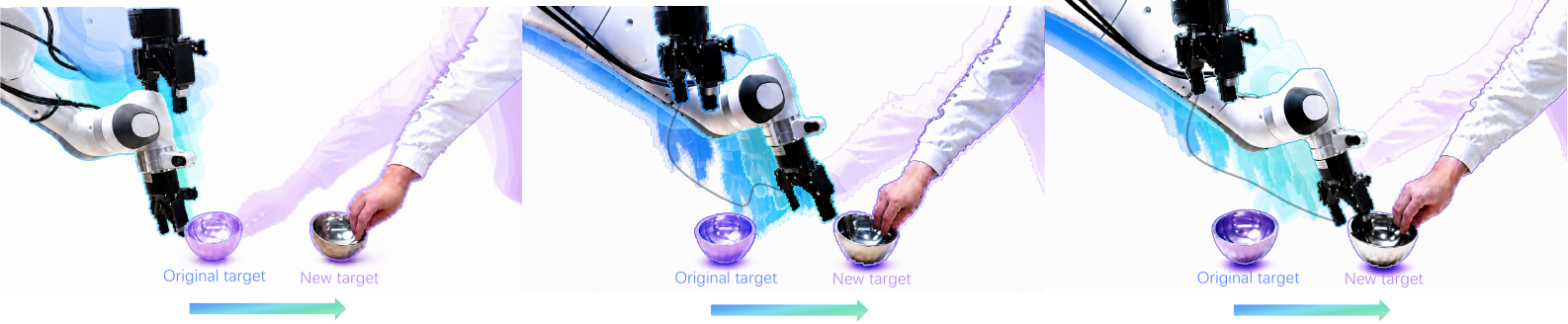}
    \caption{\textbf{Trajectory response in Bowl Stacking by Size when the bowl is suddenly moved during grasping.}
    Left: the original $\pi_{0.5}$ full-chunk policy reacts slowly because it continues executing a stale action chunk.
    Middle: the shortened action-chunk baseline updates more frequently but produces discontinuous stop-and-replan motion.
    Right: \textsc{REACT} w/o DD adapts to the moved bowl while maintaining a smoother approach trajectory.}
    \label{fig:trajectory_compare_failure}
    \vspace{-0.8em}
\end{figure}

\noindent\textbf{Long open-loop chunks remain successful and smooth, but stale.}
The full-chunk $\pi_{0.5}$ H50E50 baseline illustrates why long action chunks can look strong on standard benchmarks while failing under dynamic evaluation.
Because it executes a complete $H{=}50$ segment before replanning, it preserves strong temporal coherence, achieves the lowest average simulation jerk (658.9), and remains competitive on many standard simulated and real-robot tasks.
However, this stability comes from committing to a plan generated from an earlier observation for a long interval.
Qualitative instances of this staleness are shown in Figures~\ref{fig:pour_rice_over_pi} and~\ref{fig:trajectory_compare_failure}: in Pour Rice, H50E50 continues pouring even after the rice level exceeds the target mark; in Bowl Stacking by Size, when the bowl position is suddenly changed, it still follows the trajectory planned to grasp the original bowl before correcting.
The two specialized dynamic tasks quantify the same issue: in Pour Rice, H50E50 often keeps executing a pouring motion after the rice level has changed, yielding only $3\%$ success; in Reaction Game, its slow observation refresh dominates the response time, producing a 1505\,ms latency that trails the human teleoperation average by 800\,ms.
Thus, long open-loop chunks can preserve action-mode consistency and low jerk in standard settings, but they become stale once success depends on reacting to newly observed state changes within the chunk horizon.

\begin{figure}[tbp]
    \centering
    \includegraphics[width=0.78\linewidth]{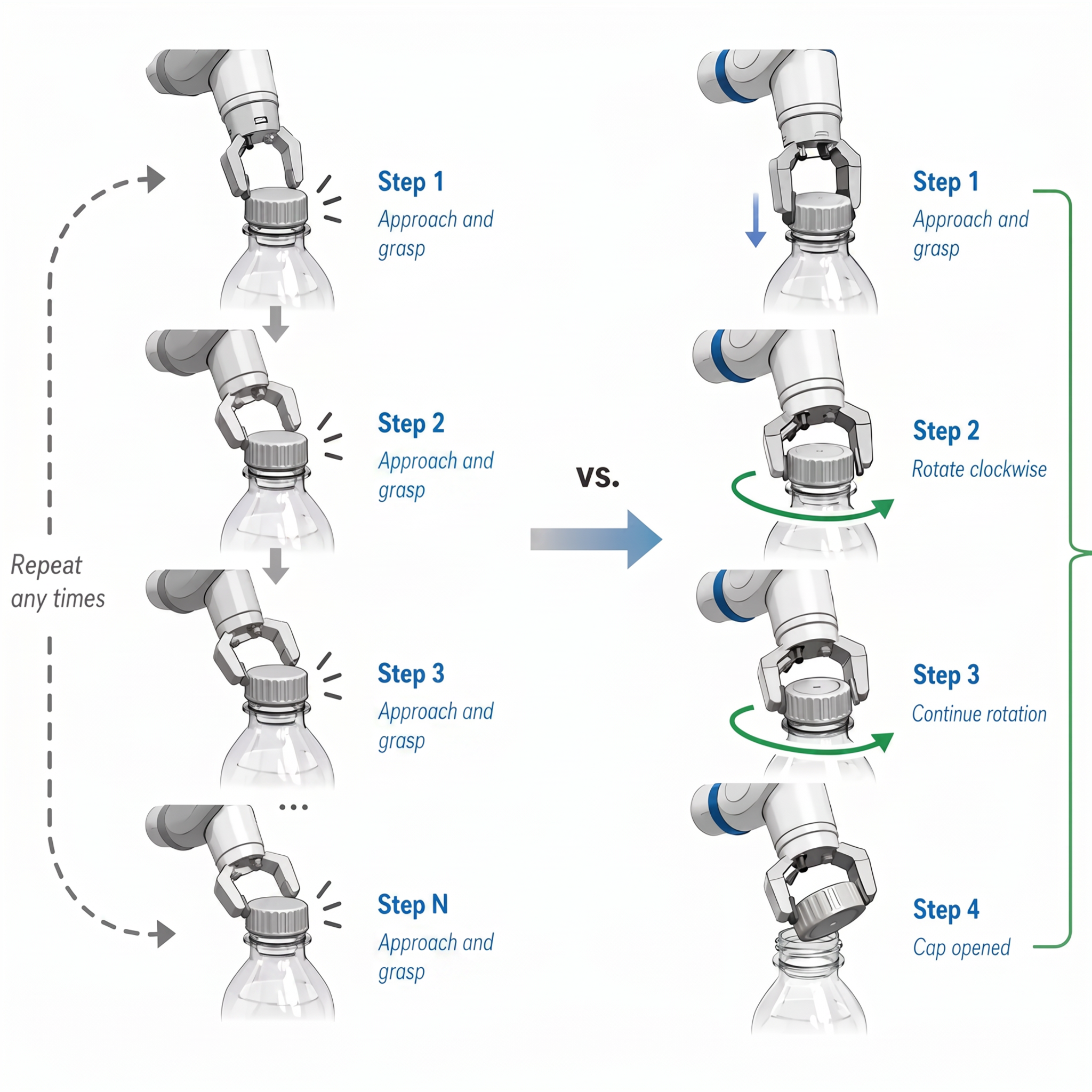}
    \caption{\textbf{Repeated cap-contact retries under shortened execution in Bottle Cap Unscrewing.}
    Representative frames from the H50E10 baseline show the policy repeatedly returning to the cap-contact region instead of sustaining the continuous grasp--twist--release sequence required to open the bottle.}
    \label{fig:screw_cap_retry}
    \vspace{-0.8em}
\end{figure}

\noindent\textbf{Shortening the execution horizon does not work.}
H50E10 attempts to improve reactivity in the most direct way: it keeps the $H{=}50$ training horizon but executes only the first 10 actions before replanning.
This shorter execution horizon fails for two reasons.
First, many contact-rich manipulation skills require an action prefix to carry the object into a new physical state before the next observation becomes informative.
When only a short prefix is executed, the robot often reaches the vicinity of contact but does not complete the state-changing part of the skill, such as closing a stable grasp, initiating a twist, or releasing after rotation.
The next replanning step therefore sees an observation that is still close to the previous local state, causing the policy to re-enter the same approach or grasping intent rather than progress to the next stage.
This produces the stop-and-replan behavior observed near contact: the arm approaches the object, pauses or retries, and becomes trapped in a narrow behavioral region.
Once such a short-prefix rollout enters an out-of-distribution contact configuration, frequent replanning further increases the chance of remaining in that region, because each new prediction is conditioned on a poorly supported local observation and again executes only a short recovery-like prefix.
The task-level results reflect this failure mode.
Compared with H50E50, H50E10 drops from $37.43\%$/$11.43\%$ to $33.71\%$/$8.71\%$ on RoboTwin Clean/Randomized settings and from $58.7\%$ to $35.9\%$ on real robots.
The degradation is especially sharp in Bottle Cap Unscrewing, where H50E10 obtains $0\%$ success on both ARX X5 and Franka.
This task requires a continuous grasp--twist--release sequence, but the shortened execution horizon repeatedly pulls the policy back into cap-grasping behavior before it can sustain the rotational phase.
Figure~\ref{fig:screw_cap_retry} illustrates this failure: the robot repeatedly returns to the cap-contact region instead of carrying the interaction through the full twisting motion.
Second, shortening the execution horizon increases the number of replanning boundaries, which amplifies multimodal action ambiguity.
Recent work on aliased robot manipulation shows that similar frame-conditioned observations can admit multiple plausible short-horizon intents, so independently replanned chunks may sample different modes across adjacent steps~\citep{lian2026intentvla}.
In H50E10, frequent replanning exposes the policy to this ambiguity at every 10-action boundary: adjacent chunks can be drawn toward different plausible short-horizon intents for nearly the same visual state, yielding inconsistent inter-chunk transitions even when each individual chunk is locally reasonable.
The middle panel of Figure~\ref{fig:trajectory_compare_failure} shows the corresponding trajectory-level symptom: after the bowl is moved, the shortened-chunk baseline updates sooner but follows a fragmented stop-and-replan path.
This loss of action-mode consistency reduces success and makes the trajectory less smooth, with average simulation jerk increasing from 658.9 for H50E50 to 1463.7 for H50E10.
Thus, simply shortening the execution horizon improves nominal observation freshness, but it does not provide the temporally extended action support needed for stable contact-rich manipulation.

\noindent\textbf{Shortening the training horizon is worse.}
H10E10 removes the mismatch between training and execution horizons, but it also removes the long-horizon action context that chunked VLAs use to coordinate multi-stage behavior and stabilize a consistent short-horizon intent.
Instead of executing prefixes of a long plan, the policy is trained from the beginning to predict only short local segments, so it never learns the long-horizon context needed to move through the same contact-rich stages that H50E10 already struggles to complete at execution time.
This makes H10E10 more likely to fall into the two failure modes above: repeated local retries near contact and inconsistent transitions across independently predicted short segments.
The resulting policy is therefore more brittle than H50E10: success collapses to $10.86\%$/$2.43\%$ on RoboTwin Clean/Randomized settings and to $21.0\%$ on real robots.
On Franka, H10E10 reaches $0\%$ on all three general tasks, and its smoothness cannot be reported for those tasks because there are no successful episodes.
Overall, shortening the training horizon does not resolve the weaknesses of H50E10; it removes the long-horizon training signal before deployment and leads to even lower success.

\noindent\textbf{Async improves throughput but not consistency.}
Async+RTC addresses the dynamic-control bottleneck from a systems perspective by hiding inference time behind action execution.
With the same $H_e{=}10$ execution interval used by the short-horizon baselines, asynchronous execution raises the nominal throughput to $\Phi_{\text{in}}{=}3.0$\,Hz and $\Phi_{\text{out}}{=}30.0$\,Hz in Table~\ref{tab:throughput_values}, compared with $2.5$\,Hz and $25.0$\,Hz for the corresponding synchronous setting.
However, this throughput gain does not remove the control problem created by short executable chunks.
Async+RTC still predicts a future chunk from an observation that may no longer match the robot and scene state when that chunk is executed, and the policy must stitch the predicted chunk onto the trajectory already being carried out.
RTC reduces this mismatch by accounting for committed actions, but it cannot fully eliminate the inconsistency between the observation used for prediction and the state reached after intervening execution.
As a result, Async+RTC often produces abrupt corrections at chunk boundaries and high-jerk motion after contact.
The smoothness results support this interpretation: Async+RTC has the highest average simulation jerk among the main methods (2022.2), well above full \textsc{REACT} (1219.0), and on ARX X5 it raises average jerk to 284.1 versus 90.1 for \textsc{REACT}.
The same inconsistency hurts task completion, with Async+RTC averaging only $17.7\%$ across real-world general tasks, far below \textsc{REACT} w/o DD ($65.7\%$) and full \textsc{REACT} ($64.8\%$).
Per-task ARX results show the characteristic pattern: Async+RTC reaches high jerk on Bowl Stacking by Size (189.3 vs. 80.3 for \textsc{REACT}) and Cable Hanging (378.8 vs. 102.7), while its success falls to $36\%$ and $3\%$ on those tasks; in Pour Rice, it also reaches $0\%$ because stitched, jerky tilting can spill rice or destabilize the flow before the target level is reached.

\noindent\textbf{Implication for \textsc{REACT}.}
These failure modes show that dynamic VLA control cannot be improved by optimizing observation freshness or throughput alone; the controller must also preserve a coherent long-horizon action context.
Short action chunks refresh observations more frequently, but they weaken the long-horizon support needed to carry contact-rich skills through state-changing phases, introduce more replanning boundaries, and amplify multimodal short-horizon ambiguity.
Async+RTC improves the systems side of the problem by hiding inference time behind execution and increasing $\Phi_{\text{in}}$/$\Phi_{\text{out}}$, but it still dispatches short executable chunks and must stitch future predictions onto an evolving robot and scene state.
\textsc{REACT} addresses both weaknesses by keeping a rolling $H{=}50$ action buffer while refining executable blocks across fresh observations.
This design lets the policy react to new scene information without repeatedly restarting from independent short-horizon prefixes, preserving the action-distribution support that stabilizes contact-rich manipulation.
The empirical pattern matches this mechanism: \textsc{REACT} w/o DD improves real-world success over H50E10, H10E10, and Async+RTC ($65.7\%$ vs. $35.9\%$, $21.0\%$, and $17.7\%$), while full \textsc{REACT} recovers camera-rate action streaming and remains far smoother than Async+RTC (simulation jerk $1219.0$ vs. $2022.2$; ARX jerk $90.1$ vs. $284.1$).
Thus, full \textsc{REACT} is not merely a faster replanning schedule; it couples fresh observations with a persistent long-horizon action buffer, which is the combination needed for responsive and coherent robot manipulation.

\section{Throughput Comparison}
\label{appendix:throughput}

We provide a systematic analysis of input and output throughput across different VLA inference schemes. For a detailed introduction to the synchronous and asynchronous inference mechanisms in action-chunking policies, see Section~3 of FASTER~\citep{lu2026faster}. Let $H$ denote the \emph{action horizon} (total actions predicted per inference), $H_e$ denote the \emph{execution horizon} (actions actually executed before the next inference), $T_{\text{VLM}}$ the VLM forward pass latency, $T_{\text{DiT}}$ the per-step DiT latency, $N$ the number of denoising steps, $T_{\text{infer}} = T_{\text{VLM}} + N \cdot T_{\text{DiT}}$ the total inference time, $f_{\text{ctrl}}$ the robot control frequency used by serialized baselines, and $f_{\text{cam}}$ the camera/action-dispatch cadence used by dual-decoupled \textsc{REACT}. For \textsc{REACT} variants with block size $S=H/K$, the observation pipeline selects one camera frame for each rolling block update, giving a selected-observation rate of $f_{\text{cam}}/S$.

\subsection{Qualitative Analysis}

\noindent\textbf{Synchronous inference.}
In synchronous mode, inference and execution are strictly sequential: the policy first completes a full inference (VLM encoding + $N$ denoising steps), then executes $H_e$ actions open-loop before acquiring the next observation. The cycle time and throughput are:
\begin{align}
    T_{\text{cycle}}^{\text{sync}} &= T_{\text{VLM}} + N \cdot T_{\text{DiT}} + \frac{H_e}{f_{\text{ctrl}}}, \\
    \Phi_{\text{in}}^{\text{sync}} &= \frac{1}{T_{\text{cycle}}^{\text{sync}}}, \qquad
    \Phi_{\text{out}}^{\text{sync}} = \frac{H_e}{T_{\text{cycle}}^{\text{sync}}}.
\end{align}

\noindent\textbf{Asynchronous inference.}
Asynchronous execution overlaps the next inference with action execution. The policy begins computing the next chunk while still executing the current one. However, each cycle still processes only one observation:
\begin{align}
    T_{\text{cycle}}^{\text{async}} &= \max\left(T_{\text{VLM}} + N \cdot T_{\text{DiT}},\; \frac{H_e}{f_{\text{ctrl}}}\right), \\
    \Phi_{\text{in}}^{\text{async}} &= \frac{1}{T_{\text{cycle}}^{\text{async}}}, \qquad
    \Phi_{\text{out}}^{\text{async}} = \frac{H_e}{T_{\text{cycle}}^{\text{async}}}.
\end{align}
When $T_{\text{infer}} < H_e / f_{\text{ctrl}}$, inference is fully hidden and $T_{\text{cycle}}^{\text{async}} = H_e / f_{\text{ctrl}}$.

\noindent\textbf{Synchronous + acceleration (e.g., quantization, distillation, fewer denoising steps).}
Acceleration methods reduce $T_{\text{infer}}$ by lowering $T_{\text{VLM}}$, $T_{\text{DiT}}$, or $N$. Let $\alpha \in (0, 1]$ be the speedup factor such that $T_{\text{infer}}' = \alpha \cdot T_{\text{infer}}$. In synchronous mode, where inference and execution are sequential, acceleration directly reduces cycle time:
\begin{align}
    T_{\text{cycle}}^{\text{accel}} &= \alpha \cdot T_{\text{infer}} + \frac{H_e}{f_{\text{ctrl}}}, \\
    \Phi_{\text{in}}^{\text{accel}} &= \frac{1}{T_{\text{cycle}}^{\text{accel}}}, \qquad
    \Phi_{\text{out}}^{\text{accel}} = \frac{H_e}{T_{\text{cycle}}^{\text{accel}}}.
\end{align}
Note that in asynchronous mode, inference time is already hidden during execution (when $T_{\text{infer}} < H_e / f_{\text{ctrl}}$), so acceleration provides no additional throughput benefit. The primary value of acceleration is therefore in synchronous settings or when $T_{\text{infer}} > H_e / f_{\text{ctrl}}$.

\noindent\textbf{\textsc{REACT} w/o DD.}
\textsc{REACT} w/o DD restructures inference so that each denoising step consumes a fresh observation and outputs $S = H/K$ actions. Without Dual Decoupling, VLM encoding, DiT denoising, and action execution are sequential:
\begin{align}
    T_{\text{cycle}}^{\text{REACT-w/o-DD}} &= T_{\text{VLM}} + T_{\text{DiT}} + \frac{S}{f_{\text{ctrl}}}, \\
    \Phi_{\text{in}}^{\text{REACT-w/o-DD}} &= \frac{1}{T_{\text{cycle}}^{\text{REACT-w/o-DD}}}, \qquad
    \Phi_{\text{out}}^{\text{REACT-w/o-DD}} = \frac{S}{T_{\text{cycle}}^{\text{REACT-w/o-DD}}}.
\end{align}
The bottleneck is typically the VLM encoding time $T_{\text{VLM}}$, which dominates the cycle.

\noindent\textbf{Full \textsc{REACT} with Dual Decoupling.}
Dual Decoupling separates \emph{sensing}, \emph{VLM encoding}, and \emph{DiT denoising} into asynchronous pipelines that run in parallel:
\begin{itemize}[nosep,leftmargin=*]
    \item \textbf{Sensing--Action decoupling}: The sensing pipeline selects one available camera frame every $S$ camera periods, i.e., at rate $f_{\text{cam}}/S$, without waiting for block execution or VLM encoding to complete.
    \item \textbf{VLM--DiT decoupling}: A pool of $M$ parallel VLM workers processes selected frames asynchronously; the DiT consumes the most recently completed embedding without waiting for any particular VLM worker.
\end{itemize}
Let $R_{\text{VLM}}$ denote the effective aggregate service rate of the VLM worker pool.
The scheduler requires $R_{\text{VLM}} \geq f_{\text{cam}}/S$.
If $M$ workers run on independent devices and each worker has latency $T_{\text{VLM}}$, then $R_{\text{VLM}}=M/T_{\text{VLM}}$.
When workers are colocated on one GPU, $R_{\text{VLM}}$ denotes the effective pool throughput under that colocated deployment.
In our real-robot \textsc{REACT} deployment, we use $M{=}2$ VLM workers colocated on a single NVIDIA RTX 4090 GPU together with the action-expert scheduler.
This evaluated layout is not a requirement of the method: \textsc{REACT} also supports multi-GPU deployment, where multiple VLM workers can be distributed across devices when additional VLM service rate or memory headroom is needed.
In the main setting, $S{=}10$ and $f_{\text{cam}}{=}30$\,Hz, so the selected-observation rate is only $3$\,Hz, well below the single-worker reference throughput of approximately $1/T_{\text{VLM}}=27$\,Hz.
Thus, the main \textsc{REACT} setting does not rely on linear multi-worker speedup; additional workers mainly provide asynchronous buffering and support smaller-block settings.
Under this provisioning condition, the input throughput is bounded by the selected observation frequency and the DiT update rate:
\begin{align}
    \Phi_{\text{in}}^{\text{REACT}} &= \min\left(\frac{f_{\text{cam}}}{S},\; \frac{1}{T_{\text{DiT}}}\right), \\
    \Phi_{\text{out}}^{\text{REACT}} &= \min\left(f_{\text{cam}},\; \frac{S}{T_{\text{DiT}}}\right).
\end{align}
When the DiT is also faster than the selected observation schedule, this simplifies to $\Phi_{\text{in}}^{\text{REACT}}=f_{\text{cam}}/S$ and $\Phi_{\text{out}}^{\text{REACT}}=f_{\text{cam}}$.
For example, with $f_{\text{cam}} = 30$\,Hz and $S{=}5$, the selected-observation rate is $6$\,Hz, which is still below the single-worker reference throughput of approximately $27$\,Hz; with $T_{\text{DiT}} = 3$\,ms, the DiT ceiling is about $333$\,Hz.
The system therefore processes selected observations at $6$\,Hz while streaming executable actions at $30$\,Hz when the worker pool sustains the selected stream.
Thus, under the same provisioning condition, output throughput reaches the camera cadence for any $S$ whenever the VLM pool and DiT sustain the selected stream.

\begin{figure}[tbp]
    \centering
    \includegraphics[width=\linewidth,trim=0 0 0 0,clip]{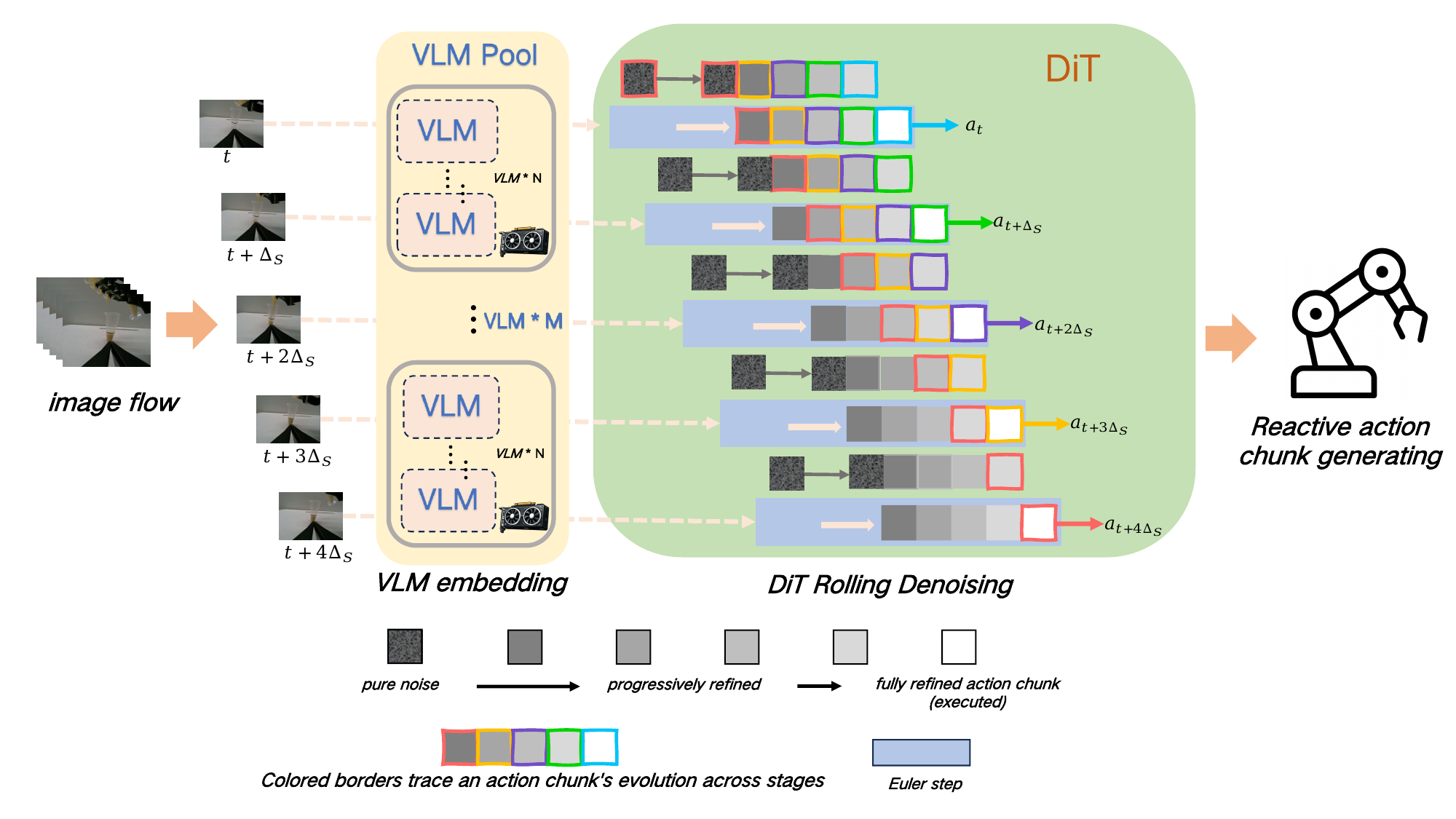}
    \vspace{-1.2em}
    \caption{\textbf{Dual Decoupling timing diagram.} Sensing, VLM encoding, DiT denoising, and action execution run as separate streams. The scheduler selects observations every $\Delta_S=S/f_{\mathrm{cam}}$ from the camera stream, and parallel VLM workers encode them into a latest-ready embedding cache; each update consumes the freshest completed embedding and the rolling action buffer, then commits the next $S$ executable actions. The architecture removes both the execution wait and the VLM wait from the DiT critical path: camera-rate input occurs in the $S=1$ setting, while output action throughput reaches the camera cadence whenever the VLM pool and DiT keep up.}
    \label{fig:dual_decoupling}
    \vspace{-0.8em}
\end{figure}

\subsection{Summary of Throughput Formulas}

Table~\ref{tab:throughput_formulas} summarizes the throughput expressions for each inference scheme.

\begin{table}[tbp]
\centering
\caption{Throughput formulas for different inference schemes. $T_{\text{infer}} = T_{\text{VLM}} + N \cdot T_{\text{DiT}}$.}
\label{tab:throughput_formulas}
\small
\begin{tabular}{lcc}
\toprule
Method & $\Phi_{\text{in}}$ (obs/sec) & $\Phi_{\text{out}}$ (act/sec) \\
\midrule
Synchronous & $\dfrac{1}{T_{\text{infer}} + H_e / f_{\text{ctrl}}}$ & $\dfrac{H_e}{T_{\text{infer}} + H_e / f_{\text{ctrl}}}$ \\[2ex]
Sync + Accel ($\alpha$) & $\dfrac{1}{\alpha T_{\text{infer}} + H_e / f_{\text{ctrl}}}$ & $\dfrac{H_e}{\alpha T_{\text{infer}} + H_e / f_{\text{ctrl}}}$ \\[2ex]
Asynchronous & $\dfrac{1}{\max(T_{\text{infer}}, H_e / f_{\text{ctrl}})}$ & $\dfrac{H_e}{\max(T_{\text{infer}}, H_e / f_{\text{ctrl}})}$ \\[2ex]
\midrule
\textsc{REACT} w/o DD & $\dfrac{1}{T_{\text{VLM}} + T_{\text{DiT}} + S / f_{\text{ctrl}}}$ & $\dfrac{S}{T_{\text{VLM}} + T_{\text{DiT}} + S / f_{\text{ctrl}}}$ \\[2ex]
\textsc{REACT} & $\min\left(\frac{f_{\text{cam}}}{S}, \frac{1}{T_{\text{DiT}}}\right)$, if $R_{\text{VLM}}\!\geq\!f_{\text{cam}}/S$ & $\min\left(f_{\text{cam}}, \frac{S}{T_{\text{DiT}}}\right)$ \\[1ex]
\bottomrule
\end{tabular}
    \vspace{-0.8em}
\end{table}

\subsection{Comparison with Typical Values}

\noindent\textbf{Measurement setup.}
We base our analysis on $\pi_{0.5}$~\cite{intelligence2025pi_}, a state-of-the-art VLA model with a PaliGemma/SigLIP VLM backbone and a flow-matching action expert. We use empirical reference latencies from a single NVIDIA RTX 4090 GPU after applying \texttt{torch.compile} optimization. The VLM forward pass (including vision encoders and language model) takes approximately $T_{\text{VLM}} = 37$\,ms; each DiT denoising step takes approximately $T_{\text{DiT}} = 3$\,ms. With $N = 10$ denoising steps, the total inference time is $T_{\text{infer}} = 37 + 10 \times 3 = 67$\,ms. The camera operates at $f_{\text{cam}} = 30$\,Hz, the robot control frequency is $f_{\text{ctrl}} = 30$\,Hz, and the action horizon is $H = 50$.

\begin{table}[tbp]
\centering
\caption{Throughput comparison based on $\pi_{0.5}$ ($H{=}50$, $f_{\text{cam}}{=}30$\,Hz, $f_{\text{ctrl}}{=}30$\,Hz, $T_{\text{VLM}}{=}37$\,ms, $T_{\text{DiT}}{=}3$\,ms, $N{=}10$).}
\label{tab:throughput_values}
\begin{tabular}{lcccc}
\toprule
Method & $H_e$ or $S$ & $T_{\text{cycle}}$ (ms) & $\Phi_{\text{in}}$ (Hz) & $\Phi_{\text{out}}$ (Hz) \\
\midrule
Synchronous & $H_e{=}50$ & $1734$ & $0.58$ & $28.8$ \\
Synchronous & $H_e{=}25$ & $900$ & $1.11$ & $27.8$ \\
Synchronous & $H_e{=}10$ & $400$ & $2.50$ & $25.0$ \\
Synchronous & $H_e{=}5$ & $234$ & $4.28$ & $21.4$ \\
Sync + Accel ($\alpha{=}0.5$) & $H_e{=}50$ & $1700$ & $0.59$ & $29.4$ \\
Sync + Accel ($\alpha{=}0.5$) & $H_e{=}25$ & $867$ & $1.15$ & $28.8$ \\
Sync + Accel ($\alpha{=}0.5$) & $H_e{=}10$ & $367$ & $2.73$ & $27.3$ \\
Sync + Accel ($\alpha{=}0.5$) & $H_e{=}5$ & $200$ & $5.00$ & $25.0$ \\
Asynchronous & $H_e{=}50$ & $1667$ & $0.60$ & $30.0$ \\
Asynchronous & $H_e{=}25$ & $833$ & $1.20$ & $30.0$ \\
Asynchronous & $H_e{=}10$ & $333$ & $3.00$ & $30.0$ \\
Asynchronous & $H_e{=}5$ & $167$ & $6.00$ & $30.0$ \\
\midrule
\textsc{REACT} w/o DD ($K{=}10$) & $S{=}5$ & $207$ & $4.84$ & $24.2$ \\
\textsc{REACT} w/o DD ($K{=}5$) & $S{=}10$ & $373$ & $2.68$ & $26.8$ \\
\textsc{REACT} (main) & $S{=}10$ & --- & $3.0$ & $30.0$ \\
\textsc{REACT} (faster input) & $S{=}5$ & --- & $6.0$ & $30.0$ \\
\textsc{REACT} (upper bound) & $S{=}1$ & --- & $30.0$ & $30.0$ \\
\bottomrule
\end{tabular}
\vspace{0.3ex}
\begin{minipage}{\linewidth}
\footnotesize
\raggedright
Throughputs are computed from unrounded cycle times; $T_{\text{cycle}}$ is rounded to the nearest millisecond for display.
For full \textsc{REACT}, $T_{\text{cycle}}$ is not listed because the serialized inference--execution cycle is removed; after the one-time warm-up, the steady-state update period is $S/f_{\text{cam}}$.
\end{minipage}
    \vspace{-0.8em}
\end{table}

\noindent\textbf{Key observations.}
\begin{itemize}[nosep,leftmargin=*]
    \item \textbf{Standard chunked VLA inference has very low input throughput.} With the full $H_e{=}50$ execution horizon, synchronous $\pi_{0.5}$ still dispatches actions at $28.8$\,Hz, close to the $30$\,Hz control cadence. However, it incorporates only $0.58$ fresh observations per second, so the policy executes a long open-loop segment before visual feedback can affect the next chunk.

    \item \textbf{Model acceleration alone does not remove the closed-loop bottleneck for long chunks.} A 2$\times$ inference speedup improves the $H_e{=}50$ synchronous input rate only from $0.58$\,Hz to $0.59$\,Hz, because the cycle time is dominated by executing the long action chunk rather than by the $67$\,ms model call. Thus faster inference is complementary, but it does not by itself make long-horizon chunked control dynamically responsive.

    \item \textbf{Shortening the action chunk improves input throughput but introduces the control-quality problems analyzed in the experiments.} Reducing $H_e$ lets the policy replan more often: for example, $H_e{=}5$ raises $\Phi_{\text{in}}$ to $4.28$\,Hz. The same serialized schedule also lowers $\Phi_{\text{out}}$ to $21.4$\,Hz and creates more chunk boundaries, matching the reduced success and higher jerk observed for short-execution baselines in \S\ref{sec:experiments}.

    \item \textbf{Asynchronous execution improves both input and output throughput, but it does not eliminate observation--state mismatch.} For $H_e{=}10$, async raises throughput from $2.50$\,Hz/$25.0$\,Hz in the synchronous setting to $3.00$\,Hz/$30.0$\,Hz by hiding inference behind execution. However, each future chunk is still predicted from an observation that may no longer match the robot and scene state at execution time, which is the mismatch analyzed for Async+RTC in the experimental results.
\end{itemize}
Together, these observations motivate \textsc{REACT}: rolling denoising preserves a long-horizon buffer with multi-observation refinement, while Dual Decoupling recovers the async-like selected-observation and action-dispatch rates shown in Table~\ref{tab:throughput_values}.

\section{Implementation Details}
\label{appendix:implementation}

\subsection{Hardware Platforms}
\label{appendix:hardware}

\begin{figure}[H]
    \centering
    \begin{minipage}[t]{0.48\linewidth}
        \centering
        \includegraphics[width=\linewidth,height=0.20\textheight,keepaspectratio]{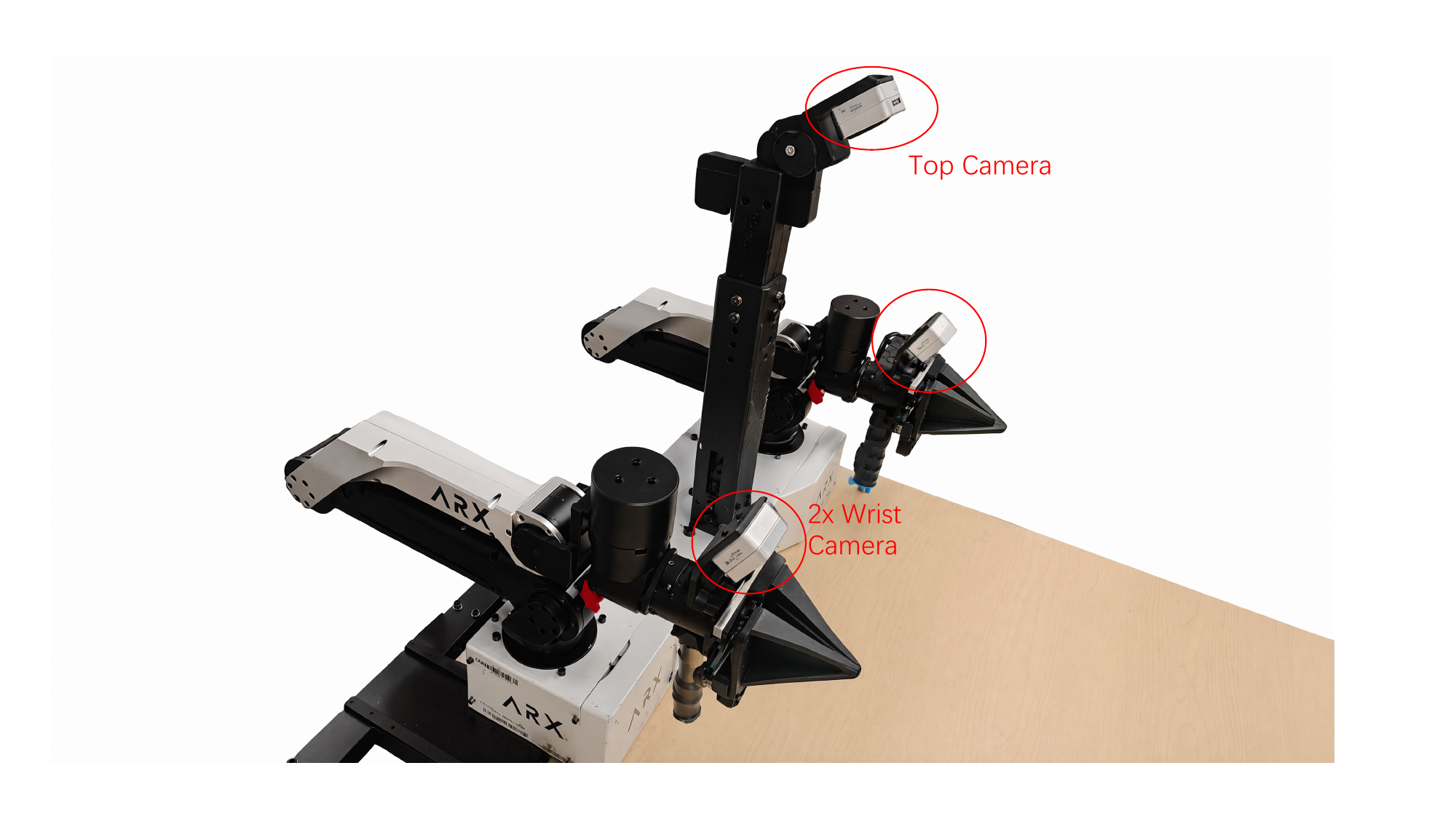}
        {\small (a) \textbf{Acone/ARX X5}}\par
    \end{minipage}
    \hfill
    \begin{minipage}[t]{0.48\linewidth}
        \centering
        \includegraphics[width=\linewidth,height=0.20\textheight,keepaspectratio]{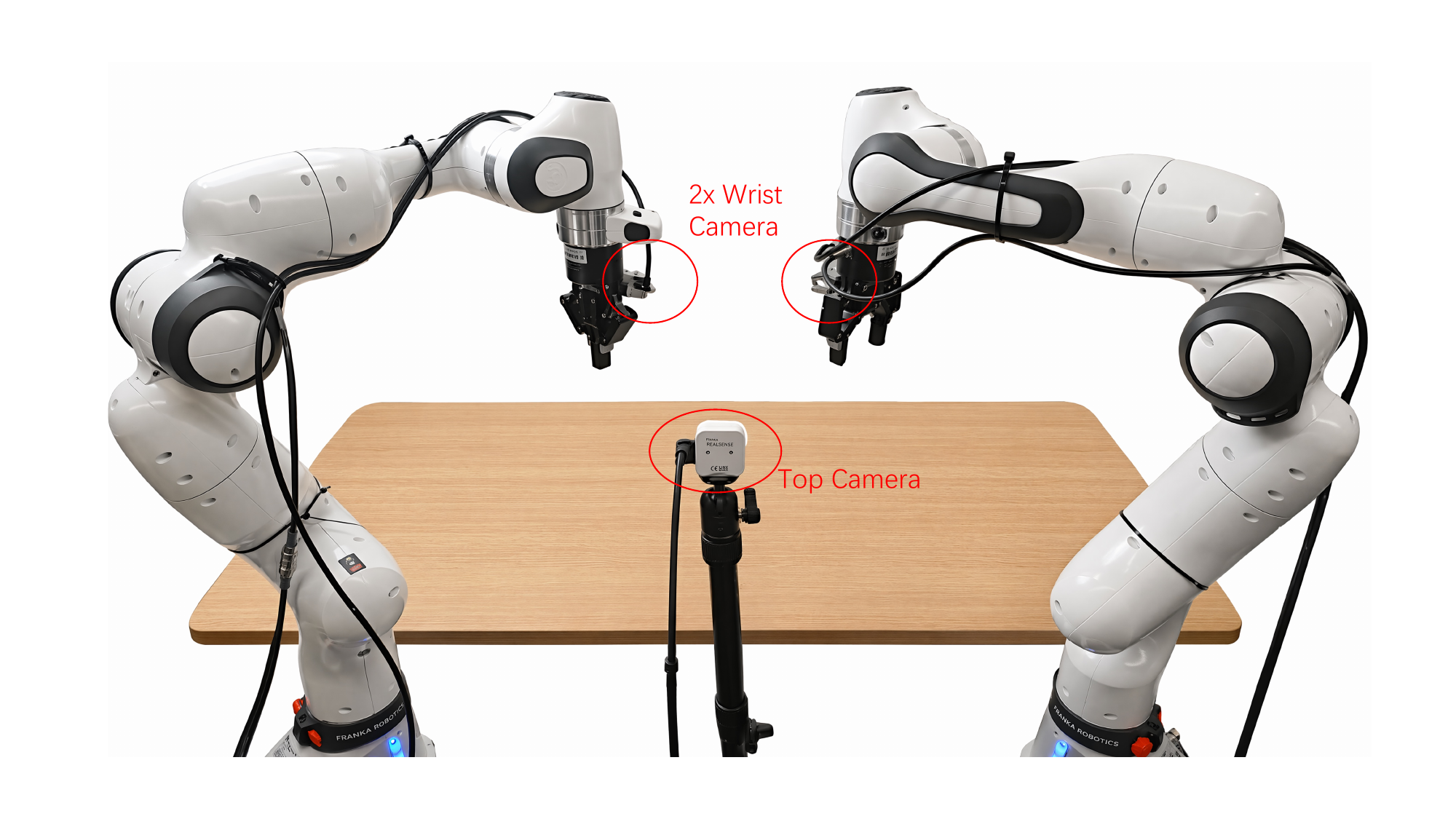}
        {\small (b) \textbf{Franka R3}}\par
    \end{minipage}
    \caption{\textbf{Hardware platforms.} Representative views of the Acone/ARX X5 bimanual platform and the Franka Research 3 dual-arm platform used in the real-robot evaluation.}
    \label{fig:hardware_platforms}
    \vspace{-0.8em}
\end{figure}

\noindent\textbf{Acone/ARX X5.}
The Acone/ARX X5 platform is a bimanual manipulation system following the ALOHA-style dual-arm setup~\citep{zhao2023learning}.
It consists of two 6-DoF robotic arms, each equipped with a 1-DoF parallel gripper.
The workspace is monitored at 30\,Hz by three RGB cameras, including a front-facing scene camera and two wrist-mounted cameras for local manipulation feedback.
The robot is controlled at 30\,Hz.

\noindent\textbf{Franka Research 3.}
We deploy a dual-arm setup comprising two Franka Research 3 (FR3) collaborative robots.
Each FR3 arm provides 7-DoF manipulation capability with torque-controlled joints and a parallel gripper.
The workspace is monitored at 30\,Hz by three RGB cameras, including a front-facing scene camera and two wrist-mounted cameras for local manipulation feedback.
Control commands are sent at 30\,Hz through the Franka Control Interface using Cartesian impedance control.

\noindent\textbf{Computational resources.}
For non-decoupled baselines and \textsc{REACT} w/o DD, inference runs on a single NVIDIA RTX 4090 GPU (24\,GB VRAM), with the VLM backbone and action expert loaded simultaneously.
For full \textsc{REACT}, our evaluated deployment also uses a single NVIDIA RTX 4090 GPU (24\,GB VRAM), colocating the action expert with $M{=}2$ asynchronous VLM workers.
This evaluated configuration keeps the same single-GPU hardware budget as the non-decoupled baselines; the scheduler only requires the effective VLM-pool service rate $R_{\mathrm{VLM}}$ to satisfy $R_{\mathrm{VLM}}\ge f_{\mathrm{cam}}/S$.
For the main setting ($S{=}10$, $f_{\mathrm{cam}}{=}30$\,Hz), the selected-observation rate is $3$\,Hz, which is well below the single-worker reference throughput of approximately $1/T_{\mathrm{VLM}}=27$\,Hz reported in Appendix~\ref{appendix:throughput}.
Thus, the evaluated single-GPU configuration satisfies the scheduler requirement without relying on linear throughput scaling from colocated workers.
\textsc{REACT} also supports multi-GPU deployment, where multiple VLM workers can be distributed across devices when additional VLM service rate or memory headroom is needed.

\subsection{Model Architecture}
\label{appendix:Architecture}

\noindent\textbf{Base VLA.}
All policy variants use $\pi_{0.5}$~\citep{intelligence2025pi_} as the base model, matching the experimental setup in \S\ref{sec:exp_setup}.
The model combines a PaliGemma-3B vision-language model with a 300M-parameter action expert.
The VLM encodes multi-view images with SigLIP and fuses them with language instructions, while the action expert generates $H$-step action chunks.
In standard $\pi_{0.5}$ inference, each generated chunk uses $N{=}10$ flow-matching denoising iterations.
We keep the backbone, tokenizer, image preprocessing, and action parameterization unchanged across baselines, \textsc{REACT} w/o DD, and full \textsc{REACT}.
The $\pi_{0.5}$ baseline schedules use the standard $N{=}10$ denoising budget per generated chunk, whereas \textsc{REACT} w/o DD and full \textsc{REACT} use the same action expert in a rolling schedule: each control update performs only one Euler denoising step on the $H$-step buffer.
An emitted \textsc{REACT} block has received multiple refinements because it persists across successive updates, not because a single \textsc{REACT} update runs a multi-step denoising loop.
\textsc{REACT} therefore changes the training timestep layout and the inference-time scheduling of the action expert rather than the model architecture.

\noindent\textbf{Action representation.}
For Acone/ARX X5, each action is represented in joint space: the two 6-DoF arm joint states plus two gripper states form a 14-dimensional vector ($12{+}2$).
For Franka R3, each action is represented in end-effector space: each arm uses Cartesian position and the continuous 6D rotation representation of Zhou et al.~\citep{zhou2019continuity}, which avoids the discontinuities of lower-dimensional 3D rotation parameterizations in neural-network regression; the two gripper states are appended, yielding a 20-dimensional vector ($2{\times}(3{+}6)+2$).
All \textsc{REACT} configurations partition the same $H{=}50$ horizon into $K$ contiguous blocks of size $S$ so that $H=KS$.

\subsection{Training Configuration}

The training hyperparameters are summarized in Table~\ref{tab:training_config}.

\begin{table}[tbp]
\centering
\caption{Training hyperparameters used for simulation and real-robot policies.}
\label{tab:training_config}
\small
\begin{tabular}{p{0.40\linewidth}p{0.52\linewidth}}
\toprule
Hyperparameter & Value \\
\midrule
Peak learning rate & $2.5 \times 10^{-5}$ \\
Final decay learning rate & $2.5 \times 10^{-6}$ \\
LR scheduler & Cosine decay with linear warmup \\
Warmup fraction & $10\%$ of the total training schedule \\
Optimizer & AdamW \\
AdamW $\beta_1$ & 0.9 \\
AdamW $\beta_2$ & 0.95 \\
AdamW $\epsilon$ & $1 \times 10^{-8}$ \\
Weight decay & $1 \times 10^{-10}$ \\
Simulation batch size & 32 \\
Simulation training length & 30,000 optimization steps \\
Real-robot training length & 20 epochs \\
Acone/ARX X5 batch size & 128 \\
Franka R3 batch size & 256 for all real-robot tasks except Reaction Game; 32 for Reaction Game \\
Gradient clipping & 1.0 \\
Mixed precision & bfloat16 \\
\bottomrule
\end{tabular}
    \vspace{-0.8em}
\end{table}

\subsection{Inference Configuration}
\label{appendix:inference_config}

\noindent\textbf{Rolling-buffer warm-up.}
At the beginning of each episode, the \textsc{REACT} buffer is initialized from the robot reset state rather than from pure noise at every position.
Let $\bar{\mathbf{s}}_0\in\mathbb{R}^{S\times D}$ denote the initial robot state tiled across one action block, and let $\tau_k=(k{+}1)/K$ be the blockwise staircase time.
The initial block at position $k$ is
\[
    B_k^{(0)}=(1-\tau_k)\bar{\mathbf{s}}_0+\tau_k\boldsymbol{\epsilon}_k,
    \qquad \boldsymbol{\epsilon}_k\sim\mathcal{N}(\mathbf{0},\mathbf{I}),
    \quad k=0,\ldots,K{-}1.
\]
Thus, for example, when $K{=}5$ the buffer begins with $0.8\bar{\mathbf{s}}_0+0.2\boldsymbol{\epsilon}$ at the front and progresses through $0.6\bar{\mathbf{s}}_0+0.4\boldsymbol{\epsilon},\ldots,$ to pure noise at the tail.
During warm-up, the policy runs the same rolling update used in steady state, but conditions the $K{-}1$ updates on the initial observation $\mathbf{o}_0=(\{I_c^0\}_{c=1}^{C},\,l,\,s_0)$: it performs one action-expert denoising pass, shifts the buffer left, and appends a fresh noise block at the tail.
The front block produced during these warm-up iterations is not used for evaluation; the controller holds its reset pose or a task-specific safe pose while the pipeline is filled.
After $K{-}1$ warm-up iterations, the next emitted block has traversed all $K$ staircase stages from the tail noise block to the executable front block.
This one-time startup cost is incurred only at episode reset; all subsequent control iterations emit one block of $S$ actions while incorporating the latest available observation.

\noindent\textbf{Dual Decoupling workers.}
For full \textsc{REACT}, VLM workers run asynchronously and publish completed visual-language embeddings to the latest-ready cache described in \S\ref{par:dual_decoupling}.
This scheduling removes the VLM pass from the DiT critical path while preserving the same policy weights as \textsc{REACT} w/o DD.

\noindent\textbf{Latency reporting.}
Table~\ref{tab:real_inference_latency} reports the synchronous per-update latency.
Because each \textsc{REACT} update performs only one action-expert denoising step, \textsc{REACT} also obtains the inference-speed benefit targeted by recent single-step or near-single-step flow-policy methods~\citep{jia2026action,lu2026faster}.

\begin{table}[H]
\centering
\caption{\textbf{Real-robot per-update inference latency.} Mean synchronous policy-update latency measured with \texttt{torch.compile} enabled.}
\label{tab:real_inference_latency}
\small
\setlength{\tabcolsep}{3pt}
\begin{tabular}{@{}lccp{0.42\linewidth}@{}}
\toprule
Policy & Latency (ms) $\downarrow$ & Speedup & Notes \\
\midrule
Standard $\pi_{0.5}$ & 66.04 & 1.00$\times$ & VLM embedding + 10 denoising steps. \\
\textsc{REACT} w/o DD & \textbf{39.96} & \textbf{1.65$\times$} & VLM embedding + single denoising step. \\
\bottomrule
\end{tabular}
    \vspace{-0.8em}
\end{table}

\subsection{Train-Inference Consistency}
\label{appendix:train_inference_consistency}

Let $k \in \{0,1,\ldots,K{-}1\}$ denote the block position in the rolling buffer, ordered from the executable front block to the far-future tail block.
The staircase training objective aligns with rolling inference by using the same time-position grid at every update and along each block's rolling trajectory.

\noindent\textbf{Per-step alignment.}
Before each \textsc{REACT} inference update, block $k$ is queried with flow time $\tau_k=(k{+}1)/K$, matching the staircase profile in Eq.~\eqref{eq:time_schedule}.
The action expert therefore receives the same relative block position and flow-time pair during deployment as it receives during training.
After one Euler update with step size $1/K$, the front block moves from $\tau_0=1/K$ to the clean endpoint and is dispatched for execution, while every later block advances to the denoising stage previously occupied by its predecessor.
Thus, every time-position pair queried during a \textsc{REACT} update corresponds to a blockwise noise level supervised by the training loss of Eq.~\eqref{eq:multi_t_loss}.

\noindent\textbf{Rolling-buffer invariant.}
The block shift preserves this alignment across time.
After the Euler update, \textsc{REACT} removes the executed front block, shifts blocks $1,\ldots,K{-}1$ one position toward the front, and appends a fresh Gaussian-noise block at the tail.
For $k=0,\ldots,K{-}2$, the shifted block that was previously at flow time $(k{+}2)/K$ now occupies position $k$ and has just been advanced by one Euler step, so its effective flow time becomes $(k{+}1)/K$.
The appended tail block is initialized as pure noise and occupies flow time $1$.
By induction, the rolling buffer recovers the same staircase profile before every inference update.

\noindent\textbf{Block-trajectory alignment.}
A block enters the buffer at position $K{-}1$ as pure noise, receives one observation-conditioned denoising update at each subsequent control iteration, and exits at position $0$ after traversing the $K$ staircase stages.
Training does not require a separate inference-specific loss for this trajectory: the staircase objective supervises all intermediate stages $\{1/K,2/K,\ldots,1\}$ within each demonstration chunk.
Consequently, every time-position query that a block receives during rolling inference has a corresponding supervised blockwise noise level in training, and the deployed scheduler does not query unseen time-position combinations.

\subsection{Benchmarks and Evaluation Protocol}
\label{appendix:benchmark_protocol}

\noindent\textbf{Simulation benchmark.}
RoboTwin 2.0~\citep{chen2025robotwin} provides a scalable bimanual manipulation benchmark with 50 complex tasks, expert demonstrations, and domain randomization for robust policy evaluation.
We use seven representative RoboTwin 2.0 tasks: \texttt{beat\_block\_hammer}, \texttt{click\_bell}, \texttt{move\_playingcard\_away}, \texttt{pick\_diverse\_bottles}, \texttt{press\_stapler}, \texttt{rotate\_qrcode}, and \texttt{turn\_switch}.
These tasks cover tool use, precision contact, grasping, card manipulation, and articulated-object manipulation.
Each policy is trained with 50 expert demonstrations per task from the official \texttt{demo\_clean} split and evaluated under two conditions: \textbf{Clean}, which uses the default scene distribution, and \textbf{Randomized}, which applies strong domain randomization.
We run 100 evaluation rollouts per task and condition following the default RoboTwin protocol.

\noindent\textbf{Real-world tasks.}
For real-world experiments, we deploy on two bimanual robot platforms: ARX Robotics X5 (ARX X5), an ALOHA-style dual-arm platform, and Franka Research 3 (Franka R3), a dual-arm setup.
The two platforms share \textbf{Bowl Stacking by Size} and \textbf{Bottle Cap Unscrewing}; ARX X5 additionally evaluates \textbf{Cable Hanging}, while Franka R3 evaluates \textbf{Keyring Hanging}.
Together, these real-world tasks test size-based reasoning, bimanual coordination, and deformable or articulated object manipulation.
Each task is tested over 30 trials, and all methods use the same randomized object placements to isolate policy differences from scene variation.

\subsection{Evaluation Metrics}
\label{appendix:evaluation_metrics}

\noindent\textbf{Success rate.}
For task-completion evaluations, success rate is the percentage of completed trials among all evaluation trials under the corresponding protocol.
The simulation benchmark uses 100 rollouts per task and condition, while each real-world general task uses 30 trials.

\noindent\textbf{Trajectory smoothness.}
Trajectory smoothness measures motion quality as the mean jerk norm of the executed joint trajectory.
Given a trajectory $\mathbf{x}(t) \in \mathbb{R}^{D}$ over $T$ timesteps, we estimate jerk $\mathbf{j}(t)$ by applying central differences three times to approximate the third derivative and report $\frac{1}{T}\sum_t\|\mathbf{j}(t)\|_2$.
Lower values indicate smoother motion.
We average this metric only over successful episodes, so comparisons are not inflated by methods that fail early or terminate before completing the task.

\noindent\textbf{Specialized-task metrics.}
Pour Rice reports task success rate on the ARX X5 platform.
Reaction Game reports mean $\pm$ standard deviation latency in milliseconds on the Franka R3 platform, measured from the visual stimulus change to the robot mouse click.
We additionally discuss excess latency over the human teleoperation average in the main text to isolate the policy-induced component of the reaction time.

\noindent\textbf{Reaction Game timing apparatus.}
We measure Reaction Game latency with a single closed-loop timing script that drives both the visual stimulus and the latency readout on the same monitor. The script controls a colored indicator displayed on the monitor and switches it from a neutral color to green at a randomized onset time $t_{\mathrm{green}}$, recorded with the system monotonic clock at the moment the green frame is committed to the display buffer. A USB keyboard sits in front of the robot, and the policy outputs end-effector actions that drive the gripper to press a designated key on this keyboard. A background keyboard listener registers the keypress timestamp $t_{\mathrm{click}}$ at the operating-system event level. The reported latency is $\Delta t = t_{\mathrm{click}} - t_{\mathrm{green}}$ in milliseconds. After the keypress is detected, the script immediately renders $\Delta t$ as a large numeric overlay on the same monitor, so each trial's measured latency is both logged to disk and visible on-screen for verification during evaluation. The human teleoperation baseline uses the identical script, monitor, keyboard, and timing path, with a human operator teleoperating the same Franka R3 arm to press the same key; this controls for monitor refresh, keyboard polling, and USB transport delays, so excess-latency comparisons against the human average isolate the policy-induced component of the reaction time.

\FloatBarrier
\section{Full Trajectory Smoothness Results}
\label{appendix:smoothness_full}

Table~\ref{tab:smoothness_full} reports the per-task trajectory smoothness values summarized in Figure~\ref{fig:smoothness_comparison}.
For reference, the same metric computed on the imitation-learning demonstrations averages 570.1 in simulation, 41.9 on ARX X5, and 15.3 on Franka; the full-execution-horizon H50E50 policy is closest to these values, while full \textsc{REACT} substantially reduces jerk relative to the real-time Async+RTC baseline.

\begin{table}[H]
\centering
\caption{\textbf{Full trajectory smoothness results.} Mean jerk norm of executed joint trajectories for each task and platform; lower indicates smoother motion. Best in \textbf{bold}, second best \underline{underlined}.}
\label{tab:smoothness_full}
\resizebox{\linewidth}{!}{%
\begin{tabular}{@{}l l cccccc@{}}
\toprule
Setting & Task & $\pi_{0.5}$ H50E50 & $\pi_{0.5}$ H50E10 & $\pi_{0.5}$ H10E10 & Async+RTC & \textsc{REACT} w/o DD & \textsc{REACT} \\
\midrule
\multirow{7}{*}{Simulation}
    & beat\_block\_hammer    & \textbf{463.4} & 996.7 & -- & 1755.0 & \underline{664.0} & 775.9 \\
    & click\_bell            & \textbf{502.4} & 1087.2 & 1341.8 & 1946.4 & \underline{690.8} & 712.0 \\
    & move\_playingcard\_away & \textbf{814.3} & 2012.7 & 1784.5 & 1832.2 & \underline{1165.8} & 1341.8 \\
    & pick\_diverse\_bottles & \textbf{857.7} & 1391.2 & -- & 1688.0 & \underline{1253.4} & 1621.8 \\
    & press\_stapler         & \textbf{567.9} & 1295.5 & 1425.7 & 3168.2 & \underline{1136.4} & 1204.7 \\
    & rotate\_qrcode         & \textbf{965.5} & 1663.9 & 1638.8 & 1944.5 & \underline{1296.1} & 1422.3 \\
    & turn\_switch           & \textbf{441.1} & 1798.8 & \underline{1153.1} & 1821.2 & 1246.7 & 1454.3 \\
\cmidrule{2-8}
& \textit{Simulation Avg.}      & \textbf{658.9} & 1463.7 & 1468.8 & 2022.2 & \underline{1064.7} & 1219.0 \\
\midrule
\multirow{3}{*}{ARX X5}
    & Bowl Stacking by Size     & \textbf{55.3} & 76.8 & 78.5 & 189.3 & \underline{70.9} & 80.3 \\
    & Bottle Cap Unscrewing     & \textbf{53.7} & -- & -- & -- & \underline{69.4} & 87.3 \\
    & Cable Hanging             & \textbf{68.9} & 95.4 & 95.0 & 378.8 & \underline{87.6} & 102.7 \\
\cmidrule{2-8}
& \textit{ARX X5 Avg.}          & \textbf{59.3} & 86.1 & 86.8 & 284.1 & \underline{76.0} & 90.1 \\
\midrule
\multirow{3}{*}{Franka}
    & Bowl Stacking by Size     & \textbf{15.1} & 27.3 & -- & 28.5 & \underline{24.3} & 27.6 \\
    & Bottle Cap Unscrewing     & \textbf{28.1} & -- & -- & 34.2 & \underline{29.8} & 30.4 \\
    & Keyring Hanging           & \textbf{12.1} & 29.7 & -- & 32.0 & \underline{25.5} & 27.7 \\
\cmidrule{2-8}
& \textit{Franka Avg.}          & \textbf{18.4} & 28.5 & -- & 31.6 & \underline{26.5} & 28.6 \\
\bottomrule
\end{tabular}%
}
    \vspace{-0.8em}
\end{table}

\FloatBarrier
\section{Ablation Studies}
\label{appendix:ablation}

We ablate the two core granularity parameters in \textsc{REACT}: the block length $S$ and the number of rolling blocks $K$. The block length $S$ controls how dynamic the controller is, because the policy ingests a new observation and emits the next action block every $S$ executed actions; smaller $S$ therefore gives more frequent observation updates and more reactive closed-loop control. The block count $K$ controls how many observations jointly shape each executed action block. A block appended as noise at the tail is refined once per rolling update and reaches execution after traversing $K$ staircase stages, so each output action block has been influenced by $K$ recent observations and has undergone $K$ denoising refinements. Larger $K$ therefore corresponds to stronger action-space convergence: more visual evidence is accumulated before an action is committed.

\begin{figure}[tbp]
    \centering
    \includegraphics[width=0.48\linewidth]{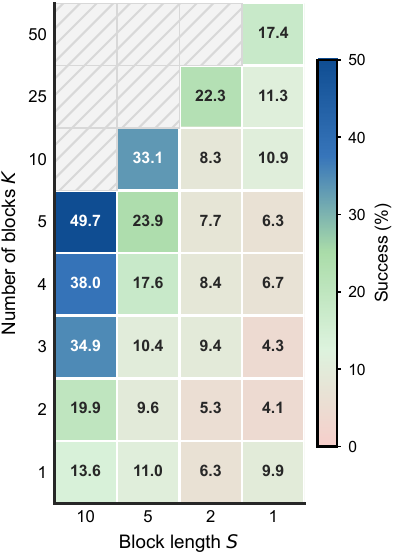}
    \caption{\textbf{Ablation on block length and block count.} Mean RoboTwin clean success rate for each $S{\times}K$ configuration; hatched cells indicate configurations not evaluated.}
    \label{fig:ablation_sk_heatmap}
    \vspace{-0.8em}
\end{figure}

\noindent\textbf{Scope of the evaluated $S{\times}K$ grid.}
The hatched cells in Figure~\ref{fig:ablation_sk_heatmap} are not arbitrary omissions; they are excluded by a backbone-level constraint of $\pi_{0.5}$. We initialize from the publicly released $\pi_{0.5}$ VLA checkpoint, which was pretrained end-to-end (VLM backbone, cross-modal conditioning, and flow-matching action expert) at a fixed action horizon of $50$. The horizon $H$ is therefore not a free hyperparameter of the action expert alone: the per-position time embeddings, the VLM--action cross-attention layout, and the action-token positional encodings are all baked in at $H{=}50$ during VLA pretraining. To inherit this pretrained VLA prior, every \textsc{REACT} configuration must satisfy $H=K\cdot S \le 50$. Consequently, configurations such as $K{=}5,S{=}5$ ($H{=}25$), $K{=}10,S{=}5$ ($H{=}50$), $K{=}1,S{\le}50$, and other $K{\times}S{\le}50$ corners are admissible, while pairs with $K\cdot S>50$ (e.g., $K{=}10,S{=}10$, $K{=}5,S{=}20$, $K{=}5,S{=}50$) would require re-pretraining the \emph{entire} VLA from scratch at a longer horizon---not only the action expert but also the cross-modal conditioning and positional encodings tied to action-token layout. Such re-pretraining would consume orders of magnitude more compute than our ablation budget and, more importantly, would conflate the effect of the rolling schedule with that of a different pretrained VLA, making the comparison non-isolating. We therefore restrict the ablation to the $K\cdot S\le 50$ region and report the densest grid within it. The same constraint explains the upper-triangular appearance of the heatmap: cells in the disallowed region cannot be evaluated under the same pretrained VLA, not because their performance is uninteresting. Within this admissible region, $S{=}10,K{=}5$ saturates the full $H{=}50$ pretrained horizon while still providing $K{=}5$ rolling refinements per emitted block, which is the largest multi-observation convergence budget the pretrained horizon allows.

\noindent\textbf{Aggregate trend.}
Figure~\ref{fig:ablation_sk_heatmap} shows both effects. Holding the block length fixed at $S{=}10$, increasing $K$ consistently improves success: the best setting $K{=}5$ reaches $49.7\%$, compared with $38.0\%$ for $K{=}4$, $34.9\%$ for $K{=}3$, and $19.9\%$ for $K{=}2$. This trend supports the role of $K$ as a convergence parameter: when each action block is denoised through more rolling stages and conditioned on more observations, the final action is more reliable. At the same time, the ablation shows that \textsc{REACT} still faces a dynamism--success trade-off. Smaller $S$ makes the policy more dynamic by refreshing observations more often, but the $S{=}1$ and other short-block configurations do not surpass $S{=}10,K{=}5$, indicating that excessive reactivity can reduce the temporal/action context available for stable execution. We therefore use $S{=}10,K{=}5$ as the final configuration because it achieves the best observed success while preserving frequent enough observation updates and sufficient multi-observation denoising for action-space convergence.

\noindent\textbf{Schedule sensitivity and practical selection.}
Success is sensitive to $(K,S)$: at fixed $S{=}10$, mean success increases from $13.6\%$ to $49.7\%$ as $K$ grows from 1 to 5, and at fixed $K{=}5$, it increases from $6.3\%$ to $49.7\%$ as $S$ grows from 1 to 10.
A larger $K$ gives each block more observation-conditioned refinements, whereas, at fixed $K$, a very small $S$ shortens the rolling horizon $H{=}KS$ and reduces action context and temporal coherence.
This suggests a simple recipe before training: choose $S$ according to the required reaction timescale and the available compute budget, since smaller $S$ requires more frequent VLM and DiT updates, and then use the largest integer $K$ satisfying $KS\le50$.
We did not tune $(K,S)$ per task or platform: the same $(K,S)=(5,10)$ setting is used in all main experiments and is optimal or tied for optimal on 6 of the 7 RoboTwin tasks, the exception being \texttt{turn\_switch}.
The single denoising update per control cycle is not an independently tuned hyperparameter: under the matched $K$-stage schedule, each block undergoes exactly $K$ refinements before execution.

\begin{figure}[tbp]
    \centering
    \includegraphics[width=\linewidth]{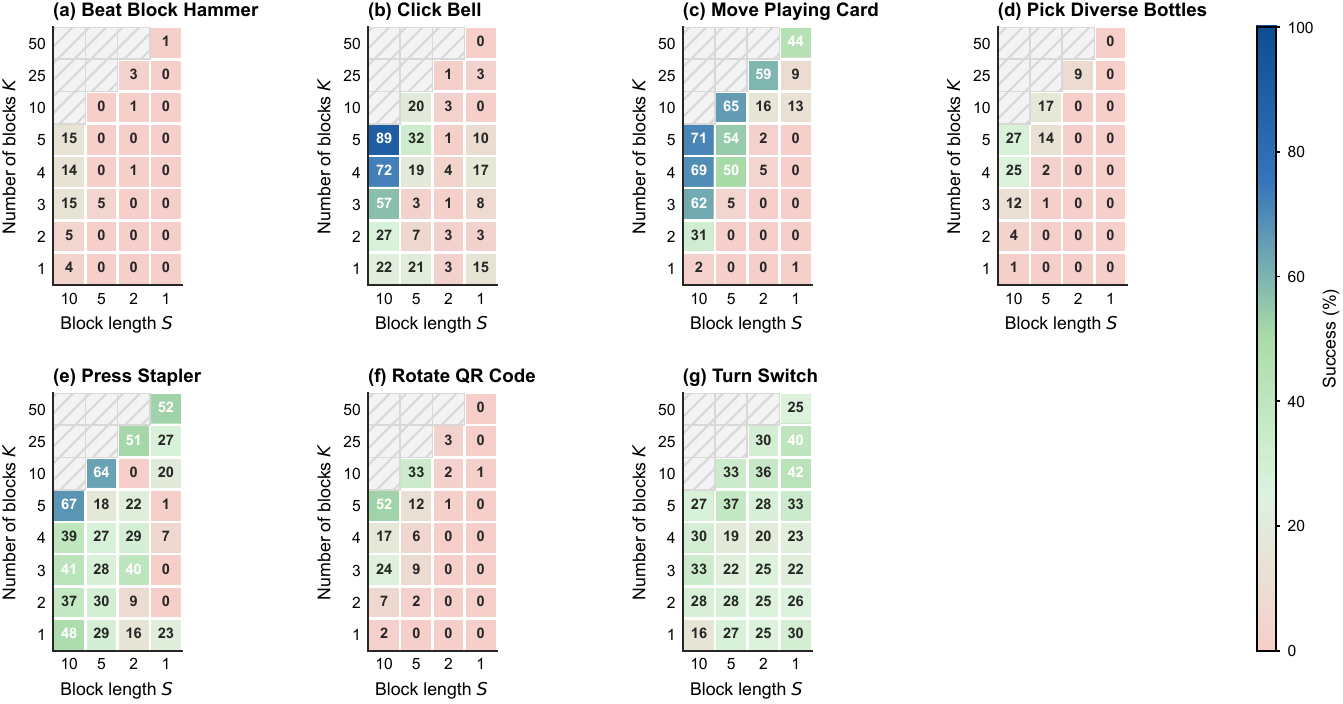}
    \caption{\textbf{Per-task ablation heatmaps.} Success rate for each of the seven RoboTwin tasks as a function of block length $S$ and block count $K$. Hatched cells indicate configurations not evaluated. The color scale is shared across all panels (0--100\%).}
    \label{fig:ablation_per_task_heatmap}
    \vspace{-0.8em}
\end{figure}

\noindent\textbf{Per-task trend.}
Figure~\ref{fig:ablation_per_task_heatmap} decomposes the mean success rate from Figure~\ref{fig:ablation_sk_heatmap} into per-task heatmaps, revealing that the best block granularity depends on the manipulation structure of each task.

\FloatBarrier
\section{OOD Dataset Action Prediction Open-Loop MSE}
\label{appendix:open_loop_analysis}

\begin{figure}[H]
    \centering
    \includegraphics[width=0.68\linewidth]{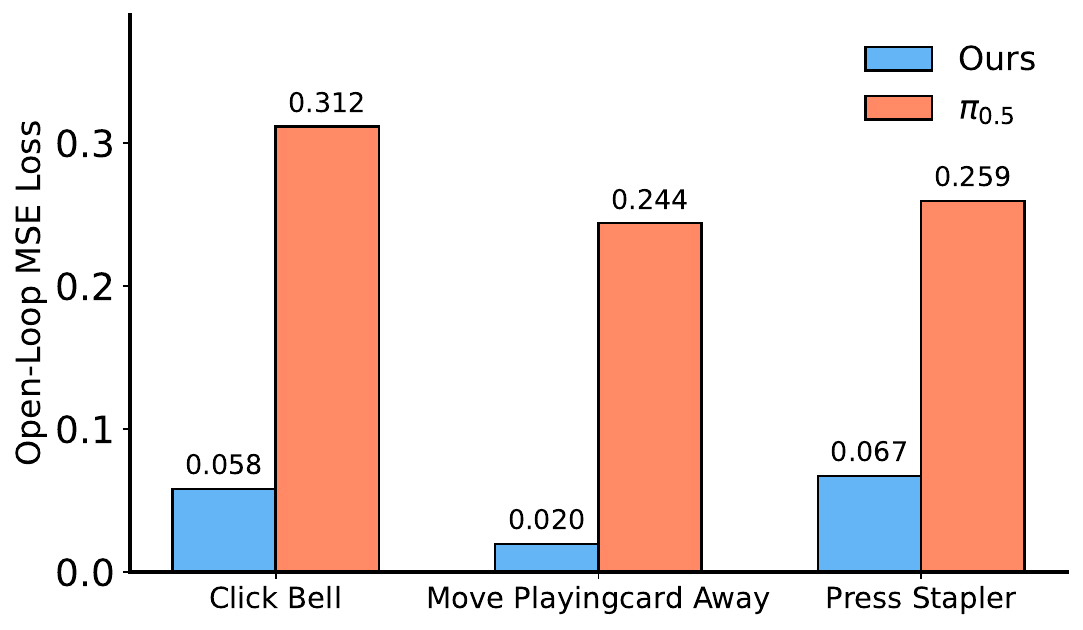}
    \caption{\textbf{OOD open-loop action prediction.} Open-loop action-prediction MSE on the \texttt{demo\_randomized} OOD dataset for three representative RoboTwin tasks.}
    \label{fig:openloop_mse}
    \vspace{-1.0em}
\end{figure}

Figure~\ref{fig:openloop_mse} provides the aggregate open-loop action-prediction diagnostic referenced in \S\ref{sec:training_efficiency}.
All open-loop MSE values and qualitative predictions in this appendix are evaluated using the final training checkpoints.

\noindent\textbf{Aggregate OOD prediction.}
Across the three \texttt{demo\_randomized} tasks, \textsc{REACT} reduces the average open-loop MSE from $0.272$ to $0.048$, an $82.4\%$ relative reduction over the $\pi_{0.5}$ baseline.
This offline result is consistent with the randomized-simulation gains in \S\ref{sec:general_results}: the staircase-trained action model fits perturbed scenes more accurately after training, rather than only changing its inference-time control schedule.

\begin{figure}[p]
    \centering
    \begin{minipage}[t]{0.48\linewidth}
        \centering
        \includegraphics[height=0.74\textheight]{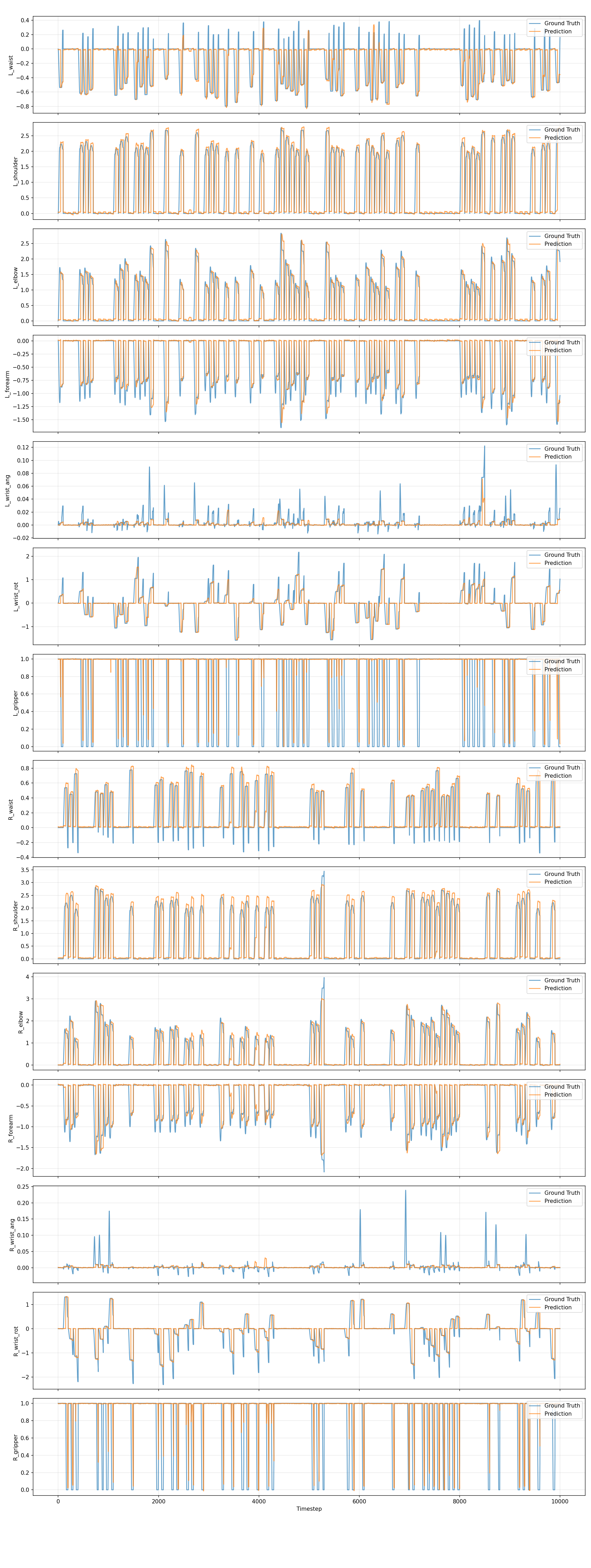}\par
        {\small (a) Baseline comparison}\par
    \end{minipage}
    \hfill
    \begin{minipage}[t]{0.48\linewidth}
        \centering
        \includegraphics[height=0.74\textheight]{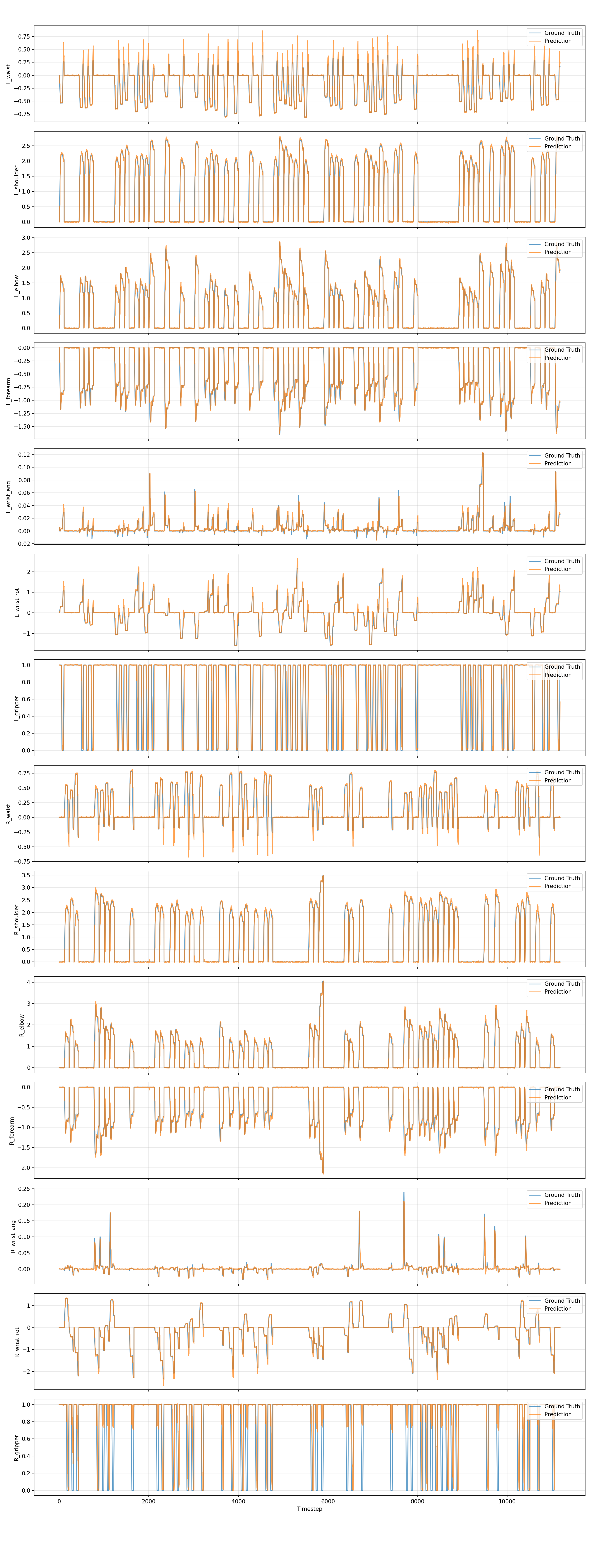}\par
        {\small (b) \textsc{REACT} comparison}\par
    \end{minipage}
    \caption{\textbf{Qualitative open-loop predictions on Move Playing Card Away.} Two visualizations of the same \texttt{demo\_randomized} example show how the predicted action sequence compares with the demonstration trajectory under the baseline and \textsc{REACT}.}
    \label{fig:openloop_qualitative}
\end{figure}

Figure~\ref{fig:openloop_qualitative} shows qualitative open-loop predictions for a representative Move Playing Card Away example from the \texttt{demo\_randomized} split.

\end{document}